%% file: 00_main.tex
\documentclass[10pt,conference]{IEEEtran}

\IEEEoverridecommandlockouts

\usepackage{amsmath,amssymb,amsfonts}%
\usepackage{amsthm}%
\usepackage{mathrsfs}%
\usepackage{bm}

\usepackage{cite}

\usepackage{url}
\usepackage{graphicx}
\usepackage{epstopdf}
\usepackage{xspace}
\usepackage{subfigure}

\usepackage{multirow}
\usepackage{booktabs} 
\usepackage{siunitx}
\usepackage{makecell}
\usepackage{makecell}

\usepackage{xcolor}
\usepackage{todonotes}
\usepackage{algpseudocode}
\usepackage{balance}
\usepackage{caption} 
\usepackage[switch]{lineno}

\usepackage[linesnumbered,ruled]{algorithm2e}

\newtheorem{theorem}{Theorem}

\theoremstyle{definition}

\newtheorem{assumption}{Assumption}

\usepackage{comment}
\usepackage[inline]{enumitem}
\usepackage[normalem]{ulem}
\usepackage{colortbl}

\begin{document}


\title{$\mathsf{RobustFSM}$: Submodular Maximization in Federated Setting with Malicious Clients}
\author{
\IEEEauthorblockN{Duc A. Tran, Dung Truong, Duy Le}
Department of Computer Science, University of Massachusetts, Boston, USA\\
Email: \{duc.tran, dung.truong001, duy.le004\}@umb.edu
}

\maketitle
\thispagestyle{plain}
\pagestyle{plain}

\begin{abstract}
Submodular maximization is an optimization problem benefiting many machine learning applications, where we seek a small subset best representing an extremely large dataset. We focus on the federated setting where the data are locally owned by decentralized clients who have their own definitions for the quality of representability. This setting requires repetitive aggregation of local information computed by the clients. While the main motivation is to respect the privacy and autonomy of the clients, the federated setting is vulnerable to client misbehaviors:  malicious clients might share fake information. An analogy is backdoor attack in conventional federated learning, but our challenge differs freshly due to the unique characteristics of submodular maximization. We propose $\mathsf{RobustFSM}$, a federated submodular maximization solution that is  robust to various practical client attacks.  Its performance  is  substantiated with an empirical evaluation study using real-world datasets.   Numerical results show that the solution quality of $\mathsf{RobustFSM}$   substantially exceeds that of  the conventional federated algorithm when attacks are severe. The degree of this improvement depends on the dataset and attack scenarios, which can  be as high as 200\%. 
\end{abstract}


\input{01_intro}
\input{02_related}

\input{03_background}

\input{04_solution}

\input{06_eval}

\input{07_conclusion}


\bibliographystyle{IEEEtran}
\bibliography{bibliography, misc,../bibfiles/tran}

\end{document}

%% file: 01_intro.tex
\section{Introduction}\label{sec:intro}
Machine learning requires data for training. Unlabeled data are abundant today but most reside at local sources. 
Usually, they would be uploaded to a central server to create samples for the training, where the labeling annotation is manually done by humans. In a resource-constrained setting, challenges arise due to the large amount of data that have to be transmitted by each local source, the server's being a processing bottleneck, and the labeling's substantial time and labor cost. Instead, a better approach is to prioritize collecting only selective data that are the most informative to the training process. This makes sense because in practice data often overlap and not every data are equally important to the training process. It is well-observed that the more data are already collected, the less marginal value new data would bring. This  is an example of a phenomenon called diminishing return, a property of submodular functions. Thus, finding the best representative subset of data above can be cast as a Submodular Maximization problem. 

\textbf{Submodular Maximization (SM). } Find a subset $S^*$ of a ground set $E$ to maximize a submodular function $f: 2^E \mapsto \mathbb{R}$ subject to some set constraint $\mathcal{I} \subset 2^E$ (for example, cardinality constraint): 
$S^* = \arg \max_S  \{  f(S): S \in \mathcal{I}  \}$. Here, $f(S)$, which is the value of subset $S$, quantifies the representativity of  $S$ with respect to ground set $E$.

\textbf{Federated SM (FSM). } In this paper, we address SM in a federated setting. The federated setting has emerged originally \cite{pmlr-v54-mcmahan17a}  in  machine learning to become a de facto standard, the Federated Learning (FL) method, for training a model using  labeled samples  distributed across many local client machines.  FSM is similar but it has a different objective function. Instead of learning a prediction model minimizing the overall prediction loss when evaluated on the  local samples of each client, FSM is learning a subset maximizing the overall representation value when evaluated using the local representation function of each client:
$S^* = \arg \max_S  \{ F(S) \triangleq \frac{1}{n} \sum_{i=1}^n  f_i(S) : S \in \mathcal{I}  \}$.
Here, $f_i(S)$ is the local submodular function representing the value of subset $S$ according to the respective criterion of client $i$. Mathematically, the key difference of FSM from FL is two-fold: the unknown to be found is a subset and the objective is a submodular set function. Whereas  FL  has great traction to date, FSM research is rare \cite{DBLP:conf/icml/Rafiey24,10286024}.

In a nutshell, the FSM architecture consists of the local machines, called the ``clients",  and a central server to coordinate the distributed computation, called the ``aggregation" server. Client $i$ has its own data and computation resource to evaluate $f_i(S)$ locally. FSM works in synchronized rounds repeatedly.   In the first step, the  server broadcasts a global solution $S$, initially randomized, to all clients. In the second step,  each client in parallel performs local SM greedy steps on its own local data to improve upon $S$, hence obtaining for itself a locally-improved solution $S_i$. In the third step, every client $i$  sends  $S_i$ to the server who in turn will perform aggregation  to obtain a new global solution. Then the first step is repeated until the global solution converges stably.

\textbf{Client risks. } The federated setting is elegant in idea, but its success relies on  clients being good citizens. Unfortunately, in FSM, bad clients can send fake local solutions to the server.  Such a solution can arbitrarily be randomized  in an ``untargeted" way to slow down the FSM process. More severely, in ``targeted"  attacks, the attacker can craft its solution in a certain way to influence the global solution toward a maliciously targeted outcome. 
  Unlike FL, for which numerous efforts have been dedicated to defending  against bad clients \cite{TranBigData2024,wang2020attack,nguyen2023backdoor}, FSM  has  emerged only recently with no research on overcoming security risks due to client misbehaviors. 

\textbf{Contributions. } In this paper, we investigate the severity of client attacks in FSM and propose a method to immunize FSM from them. 
Compared to the analogy of robust FL, our research on robust FSM poses fresh challenges.    FSM is a different type of optimization and client attacks in FSM are characteristically different from that in FL. 
We make the following key contributions.

First, considering the algorithmic process of FSM, we explore and evaluate several client attacks that are exclusive to the FSM setting. These attacks include both untargeted and targeted strategies. We have found that their impact can be substantial in terms of both convergence rate and final solution quality. Targeted attacks are more severe, which can degrade FSM performance by 90\%+ in our evaluation study. As FSM is just emerging and the degree of severity of client attacks in FSM  not well understood, our  investigation is a valuable addition to the literature of submodular maximization

Second,  we propose $\mathsf{RobustFSM}$, a robust FSM solution based on a pruning heuristic to recognize bad clients and mitigate their impact. We hypothesize that the subset of good clients should be one of the following groups: 1)  maximally-similar clients, or 2) maximally-diverse clients. It is the first subset if attacks are untargeted and the second subset if attacks are targeted. $\mathsf{RobustFSM}$ runs aggregation on both of these subset candidates, letting the clients decide which one to improve upon locally based on their own criteria. 

Third, we validate $\mathsf{RobustFSM}$ with a comprehensive evaluation  using two real-world datasets and various attack scenarios. For example, when finding top-9 images among 90K images in the PATHMNIST collection, compared to $\mathsf{FedCG}$ \cite{DBLP:conf/icml/Rafiey24} -  the conventional FSM solution, $\mathsf{RobustFSM}$ can converge faster and improve the solution quality up to 200\% when both are under the same attacks. In non-attack scenarios, our method is comparable to $\mathsf{FedCG}$.
To our knowledge, this paper is the first research effort that addresses robust SM in the federated setting. Our work can serve as a comparison benchmark for future studies.

The remainder of the paper is organized as follows. Related work is reviewed in $\S$\ref{sec:related}. Background on submodular maximization and its federated setting is provided in  $\S$\ref{sec:background}. The threat model,   pruning heuristic, and algorithmic details of $\mathsf{RobustFSM}$  are presented in $\S$\ref{sec:solution}. Evaluation results  are discussed in $\S$\ref{sec:eval}.  The paper concludes in $\S$\ref{sec:conclusion}.

%% file: 02_related.tex
\section{Related Work}\label{sec:related}
Submodularity maximization (SM) is a classic NP-hard problem in the area of combinatoric optimization.  In the early days, most SM research was focused on   monotone  functions under various constraint settings, e.g. unconstrained \cite{10.1006/jctb.2000.1989}, size-constraints  \cite{10.1007/BF01588971}, knapsack constraints \cite{10.1016/S0167-6377(03)00062-2}, and non-linear constraints \cite{10.1007/978-3-540-72792-7_15}. The celebrated result of Nemhauser et al. \cite{10.1007/BF01588971} offers a simple greedy algorithm that runs in polynomial time with optimal approximation ratio \cite{10.1145/258533.258641}. 

SM has been applied to problem solving in  numerous research areas such as machine learning, operations research, social networks, and network design. For example, SM has been useful to active learning \cite{10.5555/2208436.2208448}, feature selection \cite{10.1145/3534678.3539292}, exemplar clustering \cite{10.5555/3020948.3021001}, document summarization \cite{10.5555/2002472.2002537}, object recognition \cite{DBLP:conf/cvpr/ZhuJS14}, influence maximization \cite{10.1145/956750.956769}, stochastic optimization \cite{10.5555/2208436.2208448}, viral marketing \cite{10.1145/2487575.2487649}, social welfare  \cite{10.1145/1374376.1374389},  sensor placement \cite{8431678}, and robotic data collection \cite{pmlr-v229-akcin23a}.

Research on FSM is rare. We are aware of only two recent works \cite{DBLP:conf/icml/Rafiey24,10286024}. Rafiey \cite{DBLP:conf/icml/Rafiey24} proposed a FSM algorithm and proved its convergence properties, extending from the classic greedy method of Nemhauser et al. \cite{10.1007/BF01588971}. 
Wang et al.  \cite{10286024} addressed FSM that must protect client privacy, such that the server can still aggregate the clients' local solutions without the need to know their raw values. In contrast, our work is focused on client robustness with the goal to make FSM perform well under client attacks.

We are aware of no research addressing client misbehaviors in FSM. Although, these security risks have been addressed in a similar context,  Federated Learning (FL), in which most efforts   are to address misbehaving clients   committing backdoor attacks.  Backdoor attack is a well-known risk in machine learning. Especially for FL, due to the autonomy of clients, they can easily, stealthily, fabricate their local model to damage the global model. Also, due to the iterativity of the federated setting, an adversary can insert poison in various iterations, making this attack more challenging to counter.  There is a good research traction  on FL backdoor defense \cite{pillutla2022robust,xie2021crfl,Wu2022TowardCB,Zhang2022FLDetectorDF,10179362}.

%% file: 03_background.tex
\section{Preliminaries} \label{sec:background}

Consider a utility function $f: 2^E \mapsto \mathbb{R}$ on subsets of a given finite set $E$ (called the ground set).

\textbf{Submodular.} $f$ is   {submodular} iff for any subset $S  \subset E$ and any elements $e_1, e_2 \in E$ we have
$f(S \cup \{e_2\}) - f(S) \ge f(S \cup \{e_1, e_2\})-f(S \cup \{e_1\})$.
That is, the marginal contribution of any element  to the value of a set diminishes as the set grows. Function $f$ is said to have a diminishing return.

\textbf{Monotone. } $f$ is  {monotone} iff for any subset $S \subset E$ and any element $e \subset E$ we have
$f(S) \le f(S \cup \{e\})$.
That is, growing a set can only increase its value.

\textbf{Matroid. } Constraints on sets are often represented as a matroid.   A matroid, denoted by $M = (E, \mathcal{I})$, is a family $\mathcal{I}$ of independent sets that are subsets of a given set $E$ such that  1) $\emptyset \in \mathcal{I}$, 2) \textit{hereditary property}: if $T \in \mathcal{I}$ then $S \in \mathcal{I}$ for every $S \subset T$, and 3) \textit{augmentation property}: for any $S, T \in \mathcal{I}$ with $|S| < |T|$ then $S \cup \{e\} \in \mathcal{I}$ for some $e \in T$. For example, $\mathcal{I} = \{S \subset E: |S| \le r\}$ is a matroid representing a cardinality constraint; it is called the uniform matroid.

\subsection{Submodular Maximization (SM)} 
The problem is to find a subset $S$ of a ground set $E$ that maximizes a submodular function $f(S)$ subject to some set constraint $\mathcal{I} \subset 2^E$:
$\max_S \bigg\{  f(S): S \in \mathcal{I} \bigg\}$.
Constraint $\mathcal{I}$ usually satisfies the hereditary property. Most often, it is a matroid constraint such as the cardinality constraint, or a  knapsack constraint $\mathcal{I} = \{S \subset E : \sum_{s \in S} w_s \le b\}$.  In our paper, we assume $\mathcal{I}$ is a matroid. Denote its rank by $r \triangleq \max \{ |I| : I \in \mathcal{I}\}$.

There is a standard greedy algorithm  to solve this problem, introduced by Nemhauser et al.  \cite{10.1007/BF01588971}: start with the empty solution set $S^{(0)} = \emptyset$ and  augment it by adding a new element iteratively, $S^{(t+1)}=S^{(t)} \cup \{e^*\}$ in each step $t$, where
\[
e^*  = \arg \underset{e \in E\setminus S^{(t)}, S^{(t)} \cup \{e\} \in \mathcal{I} }{\max}  \bigg \{f(S^{(t)} \cup \{e\}-f(S^{(t)}) \bigg\} .
\]  
The iterative process stops when we cannot find any $e^*$ to add. This algorithm runs in $O(Mk)$ time, where the time unit is the time to evaluate function $f$. It achieves an approximation ratio of $(1-\frac{1}{{\mathrm{e}}})$ for the cardinality constraint (which is optimal \cite{10.1145/258533.258641}) and $\frac{1}{2}$ for a matroid constraint. 

A continuous greedy algorithm was later proposed \cite{10.1145/1374376.1374389,10.1137/080733991}, which applies to  arbitrary matroids, including cases where their space is exponentially large, with a better approximation ratio
of $(1-\frac{1}{{\mathrm{e}}})$. The idea is to convert the discrete set function to a continuous version so that we can apply a gradient descent approach to maximize it.

\textbf{Multilinear extension. } 
Let us represent a subset $S \subset E$ by a characteristic vector in $|E|$ dimensions:
\[
\mathbf{x}_S = \mathbf{x} \triangleq \bigg(\mathbf{x}[e] : e \in E \bigg) \in [0,1]^{|E|}
\]
where $\mathbf{x}[e]= 1$ if $e \in S$ and $\mathbf{x}[e]=0$ otherwise. For example, $\mathbf{x}_\emptyset = \mathbf{0}$ (all elements are zero) and $\mathbf{x}_E = \mathbf{1}$ (all elements are 1). Conversely,  an arbitrary characteristic vector $\mathbf{x}$ corresponds to a random set $S$ such that the probability of seeing $e \in S$ is $\mathbf{x}[e]$. 
The set function $f$  has a multilinear extension, $F: [0,1]^{|E|} \rightarrow \mathbb{R}$, as the following continuous function:
\begin{align}
F(\mathbf{x}) \triangleq \mathbb{E}_{S \sim \mathbf{x}} [f(S)] = \sum_{S \subset E} f(S) \prod_{e \in S} \mathbf{x}[e]\prod_{e \not \in S} (1-\mathbf{x}[e])
\end{align}
Here, $\mathbb{E}_{S \sim \mathbf{x}}$ denotes the expectation over every subset $S$ randomized from characteristic vector $\mathbf{x}$. With this definition, we have $f(S)=F(\mathbf{x}_S)$, and so intuitively the maximization of $f$ in the combinatoric space can be transformed to that of $F$ in the continuous space. Consequently, we can apply greedy gradient descent in place of the greedy forward step of the standard greedy algorithm. 
The gradient vector of the extension function $F$ is
\begin{align}
\nabla F(\mathbf{x}) = \bigg( \frac{\partial F}{\partial \mathbf{x}(e)}: e \in E \bigg)\\
\text{where~} \frac{\partial F}{\partial \mathbf{x}(e)} = \mathbb{E}_{S \sim \mathbf{x}}[ f(S \cup \{e\} - f(S \setminus \{e\})].
\end{align}
Thus,   $\nabla F(\mathbf{x})$ can be computed based on a sampling of $f(S)$. 

\textbf{Continuous greedy.} The algorithm works as follows. Start with the initial characteristic vector $\mathbf{x}^{(0)} = \mathbf{0}$ and then iterate $t=1,2,...$ to adjust it incrementally as follows
\begin{align}
&\mathbf{w}^{(t)} = \arg \max_{\mathbf{w} \in \mathcal{P}}   \langle \mathbf{w}, \nabla F(\mathbf{x}^{(t)}) \rangle  \\
&\mathbf{x}^{(t+1)} = \mathbf{x}^{(t)} + \eta  \mathbf{w}^{(t)}.
\end{align}
Here, $\eta \in (0,1)$ is the learning rate. Vector $\mathbf{w}^{(t)}$ is needed to ensure that each gradient update results in a new vector that stays in the matroid polytope $\mathcal{P}$. The iterative process stops when $\mathbf{x}^{(t+1)} \not \in \mathcal{P}$. The final solution $S$ is a random subset from characteristic vector $\mathbf{x}^{(t)}$. In the continuous version, since we seek a continuous vector $\mathbf{x}$ instead of a set $S$ subject to constraint $S \in \mathcal{I}$, we need an equivalent constraint  $\mathcal{P}$ on $\mathbf{x}$. This constraint is called a matroid polytope, which is the convex hull of the indicator vectors of the bases of matroid $\mathcal{I}$:
\[
\mathcal{P} = \{\mathbf{x} \ge \mathbf{0}: \sum_{e \in S} x[e] \le r_{\mathcal{I}}(S) ~\forall S \subset E \}
\]
where $r_{\mathcal{I}}(S) \triangleq  \max   \{ |I| : I \subset S, I \in \mathcal{I}   \}$ denotes the rank function of $S$ with respect to matroid $\mathcal{I}$. For example, if $\mathcal{I}$ is the uniform matroid, i.e., cardinality constraint  $\mathcal{I} = \{ S \subset E: |S| \le r\}$, we have $r_{\mathcal{I}}(S) = \min(|S|,r)$ and $\mathcal{P} = \{\mathbf{x} \ge \mathbf{0}: \sum_{e \in S} x[e] \le \min(|S|,r) ~\forall S \subset E\}$ $=$ $\{\mathbf{x} \in [\mathbf{0}, \mathbf{1}]: \sum_{e \in V} x[e] \le r \}$. With this particular constraint, applying the continuous greedy algorithm, once we converge with the final $\mathbf{x}$, the corresponding subset solution is the $r$-cardinality subset with the largest $r$ coordinates of $\mathbf{x}$.

\subsection{Federated Submodular Maximization (FSM)}
We assume a typical federated setting having a set $C$ of clients, where each client $i \in C$ has a local submodular function $f_i(S)$ to value a candidate subset $S$. The objective is to find a subset of $E$ under constraint $\mathcal{I}$ that maximizes the average value of these local functions, i.e.,
\[
S^* = \arg \max_S \bigg\{ f(S) \triangleq \frac{1}{|C|} \sum_{i \in C}  f_i(S) : S \in \mathcal{I} \bigg\}.
\]
Objective function $f(S)$ is submodular because it is the average of submodular functions. We assume that  ground set $E$ and constraint $\mathcal{I} \subset E$ are globally known, but function $f_i$ is private to client $i$. Clients do not communicate with one another. The server is the only coordinator. 

\textbf{Federated Continuous Greedy ($\mathsf{FedCG}$). } The continuous greedy algorithm for SM is extended for the FSM setting as follows \cite{DBLP:conf/icml/Rafiey24}. Let $F$ and $F_i$ be the multilinear extensions of $f$ and $f_i$, respectively. 
$\mathsf{FedCG}$  runs repetitively $T$ rounds, called ``aggregation rounds" (a.k.a. ``communication" or ``global" rounds in the literature). In each round $t$, the algorithm keeps track of a global solution  $\mathbf{x}^{(t)}$ and a local solution $\mathbf{x}_i^{(t)}$ for each client $i$. At initialization, $\mathbf{x}^{(0)}  = \mathbf{x}_i^{(0)} = \mathbf{0}$.
At the beginning of round $t=1,2,...$,   the server broadcasts   global solution $\mathbf{x}^{(t-1)}$ to a subset $C^{(t)} \subset C$ of  clients. Upon receipt, each client $i \in C^{(t)}$ runs one continuous greedy step to improve $\mathbf{x}^{(t-1)}$ using its local function $F_i$. As a result,
\begin{align}
&\mathbf{w}_i^{(t)} = \arg \max_{\mathbf{w} \in \mathcal{P}}   \langle \mathbf{w}, \nabla F_i(\mathbf{x}^{(t-1)}) \rangle  \\
&\mathbf{x}_i^{(t)} = \mathbf{x}^{(t-1)} + \eta  \mathbf{w}_i^{(t)}.
\label{eq:fedCG_localsolution_update}
\end{align}
Then client $i$ uploads its (polytope-projected) gradient vector $\mathbf{w}_i^{(t)}$ to the server. Having received these vectors from all clients, the server updates the global solution as follows: 
\begin{align}
\mathbf{w}^{(t)} &= \mathsf{mean} \bigg(  \{\mathbf{w}_i^{(t)}: i \in C^{(t)} \} \bigg)\\
\mathbf{x}^{(t)}  &= \mathbf{x}^{(t-1)} +   \eta \mathbf{w}^{(t)}.
\end{align}
The same process repeats for round $(t+1)$.

The reason for each client $i$ to upload to the server the projected gradient vector instead of the original gradient vector is because the former is a sparse vector hence more efficient in communication. It must be in  polytope $\mathcal{P}$ which has rank $r$, thus having all but a most $r$ coordinates equal to zero. In contrast, the original gradient vector may contain many non-zero coordinates. Hereafter, unless otherwise explicitly stated, the term ``gradient vector" refers to the projected vector.

Note that  in any round the global solution is always the average of the local solutions of $C^{(t)}$, which is  because
\begin{align*}
\mathbf{x}^{(t)}  &= \mathbf{x}^{(t-1)} +   \eta \cdot \frac{ 1}{|C^{(t)}|} \sum_{i \in C^{(t)}}    \mathbf{w}_i^{(t)} \nonumber\\
&=  \frac{ 1}{|C^{(t)}|} \sum_{i \in C^{(t)}}  \bigg(\mathbf{x}^{(t-1)} + \eta \mathbf{w}_i^{(t)} \bigg)
=  \frac{ 1}{|C^{(t)}|} \sum_{i \in C^{(t)}} \mathbf{x}_i^{(t)}.
\end{align*}

%% file: 04_solution.tex
\section{Robust FSM}\label{sec:solution}
Albeit elegant, FSM is vulnerable to   misbehaviors by the clients.   Below, we present ways a client can attack FSM. Then we propose a robust FSM algorithm  to mitigate them.

\subsection{Threat Model}\label{sec:threatmodel}
We assume a Byzantine environment where an adversary controls a fraction $\beta < 50\%$ of clients. 
The adversary knows the server’s aggregation logic and the uncompromised clients’ computation logic, 
but cannot modify any of these processes. Since local functions, $f_i$’s, of these clients are private, 
they are unbeknownst to the adversary. Hereafter, we refer to the uncompromised and compromised clients 
as good and bad clients, respectively. A bad client $i$ can misbehave arbitrarily by uploading a wrong 
version of the gradient, $\widehat{\mathbf{w}}_i^{(t)} \neq \mathbf{w}_i^{(t)}$. This version can be 
randomized or purposely fabricated. The latter case is to make the final global outcome 
$\widehat{\mathbf{x}}^{(T)}$ maximally deviate from the authentic outcome $\mathbf{x}^{(T)}$ 
or selfishly favor or disfavor certain elements $\{ e_1, e_2, \ldots \} \subset E$.

Due to the logic of the algorithm, every local gradient $\mathbf{w}_i^{(t)}$
must be constrained by the polytope $\mathcal{P}$. Therefore, a valid value 
$\mathbf{w}_i^{(t)} = \mathbf{w}$ needs $\mathbf{w} \in [0,1]^{|E|} \wedge 
\sum_{e \in E} \mathbf{w}[e] \le r$. As such, if the adversary uploads a 
fake vector $\widehat{\mathbf{w}}_i^{(t)}$ that violates this condition, 
the server can detect it. Therefore, we assume that 
$\widehat{\mathbf{w}}_i^{(t)}$ satisfies this condition. We consider the 
following most likely attacks. In practice, the adversary would apply any 
single one or combination of them on each bad client in a Byzantine manner.

\textbf{Random attack:} $\widehat{\mathbf{w}}_i^{(t)}$ is an arbitrarily random vector 
in the polytope $\mathcal{P}$. This attack introduces noise to the aggregation, 
hoping to slow down the convergence.

\textbf{Reverse attack:} This attack aims to maximally reverse the gradient of the 
global solution. Knowing the current global solution $\mathbf{x}^{(t-1)}$, 
the adversary chooses $\widehat{\mathbf{w}}_i^{(t)}$ such that its new local solution 
$\widehat{\mathbf{x}}_i^{(t)} = \mathbf{x}^{(t-1)} + \eta \widehat{\mathbf{w}}_i^{(t)}$ 
maximally reverses the preference ranking of the coordinates in $\mathbf{x}^{(t-1)}$. 
Intuitively, the more preferred an element $e \in E$ is in the global solution 
(higher value of $\mathbf{x}^{(t-1)}[e]$), the less preferred the adversary
would make it in the local solution (lower value of 
$\widehat{\mathbf{x}}_i^{(t)}[e]$). In other words, the adversary would aim for

\[
\min_{\mathbf{x}} \langle \mathbf{x}^{(t-1)}, \mathbf{x} \rangle 
= \min_{\mathbf{w}} \langle \mathbf{x}^{(t-1)}, \mathbf{x}^{(t-1)} + \mathbf{w} \rangle.
\]
Therefore, the adversary would set

\[
\begin{aligned}
\widehat{\mathbf{w}}_i^{(t)} 
&= \arg\min_{\mathbf{w} \in \mathcal{P}} 
    \langle \mathbf{x}^{(t-1)}, \mathbf{x}^{(t-1)} + \mathbf{w} \rangle \\
&= \arg\min_{\mathbf{w} \in \mathcal{P}} 
    \langle \mathbf{x}^{(t-1)}, \mathbf{w} \rangle.
\end{aligned}
\]

\textbf{Include attack:} 
This attack prioritizes certain elements 
$E_0 = \{ e_1, e_2, \ldots \} \subset E$ 
in the global solution. The adversary would maximize 
$\sum_{e \in E_0} \widehat{\mathbf{x}}_i^{(t)}[e]$ 
and in the remaining degrees of freedom minimize the ranking 
of the remaining elements; i.e., maximizing
\[
\begin{aligned}
\sum_{e \in E_0} \mathbf{x}[e] - \sum_{e \in E \setminus E_0} \mathbf{x}[e]
\\= \sum_{e \in E_0} 
\left( 
\mathbf{x}^{(t-1)}[e] + \mathbf{w}[e] 
\right)
- 
\sum_{e \in E \setminus E_0} 
\left( 
\mathbf{x}^{(t-1)}[e] + \mathbf{w}[e] 
\right).
\end{aligned}
\]
Therefore, the adversary would set
\begin{equation}
    \widehat{\mathbf{w}}_i^{(t)} 
= \arg\max_{\mathbf{w} \in \mathcal{P}} 
\left\{
\sum_{e \in E_0} \mathbf{w}[e] 
- \sum_{e \in E \setminus E_0} \mathbf{w}[e]
\right\}.
\end{equation}

The solution to (10) definitely has $\widehat{\mathbf{w}}[e] = 0$ 
for all $e \in E \setminus E_0$. Therefore, the adversary only needs 
to maximize $\sum_{e \in E_0} \mathbf{w}[e]$ considering only 
the elements in $E_0$. The maximum is reached when 
$\sum_{e \in E_0} \mathbf{w}[e] = \min(|E_0|, r)$ 
($r$ is the rank of matroid $\mathcal{I}$). 
For the cardinality constraint, one way to achieve this is 
to set $\widehat{\mathbf{w}}[e] = \min(1, \tfrac{r}{|E_0|})$ 
for all $e \in E_0$. Another way is to
\[
\max_{\mathbf{w} \in \mathcal{P}} 
\left\{
\min_{e \in E_0} 
\left( 
\mathbf{x}^{(t-1)}[e] + \mathbf{w}[e]
\right)
: 
\sum_{e \in E_0} \mathbf{w}[e] = \min(|E_0|, r)
\right\}.
\]
This is achieved when, for every $e \in E_0$,
\[
\mathbf{x}^{(t-1)}[e] + \mathbf{w}[e] 
= \min
\left( 
1, 
\frac{r}{|E_0|}
\right)
+ 
\frac{1}{|E_0|}
\sum_{j \in E_0} \mathbf{x}^{(t-1)}[j].
\]
Thus, the adversary would set, for every $e \in E_0$,
\[
\widehat{\mathbf{w}}[e] 
= \min
\left( 
1, 
\frac{r}{|E_0|}
\right)
+ 
\frac{1}{|E_0|}
\sum_{j \in E_0} \mathbf{x}^{(t-1)}[j]
- 
\mathbf{x}^{(t-1)}[e].
\]

\textbf{Exclude attack:} 
This attack aims to exclude certain elements 
$E_0 = \{ e_1, e_2, \ldots \} \subset E$ 
from the global solution. So the adversary would compute the gradient vector as usual but after that set the coordinate corresponding to each element 
$e \in E_0$ to zero; i.e.,
\[
\widehat{\mathbf{w}}_i^{(t)} 
= \arg\max_{\mathbf{w} \in \mathcal{P}} 
\langle \mathbf{w}, \nabla F_i(\mathbf{x}^{(t-1)}) \rangle,
\]
\[
\widehat{\mathbf{w}}_i^{(t)}[e] = 0, \quad \forall e \in E_0.
\]

\subsection{ Robust  Aggregation}
  It is hard to build a Byzantine-proof defense for large-scale federated computing.  Instead, in federated systems expecting performance to be good, not necessarily precise, e.g., serving machine learning tasks, a   practical approach against Byzantine attacks is to mitigate their impact. Robust  aggregation methods vary, mostly adopting heuristics \cite{Yang2024,9721118,10.1145/3154503}. 
 
 \textbf{Limitation of common heuristics. }
 Common strategies include 1) removing suspiciously bad data/clients (e.g., \cite{Yang2024}) from the aggregation, and/or 2) applying a robust aggregator such as Geometric Median (GM) (e.g., \cite{9721118}). In particular, for federated learning,   Pillutla et al.  \cite{9721118}  recommend the use of  GM instead of the conventional arithmetic mean to average models at the server.   This is no surprise because  GM  is known to have a great robustness property  as an aggregator. It has the optimal breakdown point of 1/2 \cite{huber1981breakdown}, meaning that up to half of the points in $V$  may arbitrarily be  corrupted and the geometric median remains a robust (bounded) estimator for  the location of the uncorrupted data. In contrast, the breakdown point of arithmetic mean is 0, meaning that the adversary only needs to corrupt one point to achieve a target (bad) average value.  For this reason, GM has been adopted as a robust mean approximation in many adversarial applications.
  
In our investigation, we ask ``is GM also a robust aggregator  for FSM?" At first sight, it seems so as above explained, but interestingly we have found evidence that GM is actually not; it is much worse than arithmetic mean under the Include attack (supporting results are in Section \ref{sec:eval}). 
Regarding the client filtering strategy, which removes suspiciously bad clients from the aggregation, the common heuristic is to aggregate only those clients that maintain a high level of similarity. The rationale is that the local solutions of good clients, if they are the majority, should be closer to the final solution (consensus) than that of the bad clients, and so the former should form a highly-similar cluster. This similarity heuristic is effective for  federated learning  \cite{TranBigData2024}, but not for FSM. In FSM, the heuristic could be robust for untargeted attacks such as the $\mathsf{Random}$ attack, but under highly targeted attacks such as the $\mathsf{Include}$ attack, the local solutions of bad clients should be more clustered than that of the good. 

\textbf{New heuristics for robust aggregation. } We propose $\mathsf{RobustFSM}$, a Byzantine-robust variant of $\mathsf{FedCG}$.   
For now, assume the full participation of all the clients in each round, i.e., $C^{(t)}=C$ for all $t=1, 2, ...$.  Focus on one round (the current round) and let $G\subset C$ and $B=C\setminus G$ denote the subset of good clients and the subset of bad clients in this round, respectively. Consider a maximally-similar subset $C^+_{sim} \subset C$ and a maximally-diverse subset $C^+_{div} \subset C$, both of size $q|C|= \frac{2}{3} |C| $. 
Thanks to the majority of the clients being good, we hypothesize that   one of these two subsets is dominated by the good set $B$ and is better than  the original client set $C$ to represent  $B$. The fraction $\frac{1}{3}$ of the clients excluded should mostly consist of bad clients.  Furthermore, if we know that   attacks are untargeted, such as the $\mathsf{Random}$ attack,  we only need $C^+_{sim}$. On the other hand, if we know that attacks are highly targeted, such as the $\mathsf{Include}$ attack, we only need $C^+_{div}$. However, since the type of attack is unbeknownst, the server does not know which subset is the ``good" coreset. It thus will consider both subsets, which we will hereafter refer to as   candidate coresets, compute their corresponding aggregation, and let the good clients to decide which aggregation version to use for the next round. 

\subsubsection{Good-Set Representation} 
Similarity and diversity are quantified based on gradient-pairwise distance.  Because gradients are sparse binary vectors with at most $r$ 1's, we suggest using the Hamming distance. Let us ignore the round index $t$ and denote  by $\mathbf{w}_i$ the local gradient from client $i$ in the current round.  We solve:
\begin{align}
C^+_{sim} &= \arg \min_{X \subset C, |X|=q|C|} \sum_{i \in X} \sum_{j \in X}  \mathsf{Hamming}(\mathbf{w}_i, \mathbf{w}_j)  \label{eq:Csim}\\
C^+_{div} &= \arg \max_{X \subset C, |X|=q|C|} \sum_{i \in X} \sum_{j \in X}  \mathsf{Hamming}(\mathbf{w}_i, \mathbf{w}_j)\label{eq:Cdiv}
\end{align}
Eq \eqref{eq:Csim} is equivalent to
\begin{align}
C^+_{sim} &= \arg \min_{X \subset C, |X|=q|C|} \sum_{i \in X} \sum_{j \in X} \underbrace{\mathsf{Hamming}(\mathbf{w}_i, \mathbf{w}_j)}_{\text{denoted by~} {2r}(1-c_{ij})} \nonumber\\
&= \arg \max_{X \subset C, |X|=q|C|}  \sum_{i \in X} \sum_{j \in X} c_{ij}.
\label{eq:C}
\end{align}
Thus,   optimizations   \eqref{eq:Csim} and  \eqref{eq:Cdiv} are  instances of the well-known Diversity Maximization  problem. This problem is NP-hard  but effective heuristic algorithms exist \cite{MDP}. We adopt the classic greedy algorithm by Glover et al. \cite{MDP}. To obtain $C^+ \triangleq C^+_{sim}$ (similarly for $C^+_{div}$), the algorithm works simply as follows. Start with $C^+ = \emptyset$, then repeat adding  one client at a time that is furthermost from  $C^+$, 
\begin{align*}
i^* = \arg \max_{ i  \in C \setminus C^+} \sum_{j \in C^+} c_{ij},~C^+ = C^+ \cup \{ i^*\}
\end{align*}
until $|C^+|  = q|C|$. 
For convenience, denote  this algorithm by
$X^+= \mathsf{coreset}_{sim | div}(X; q)$ 
 where $X$ is the set of input gradients,  $q = |X^+|/|X|$ is the  proportion ratio for the cardinality of the output coreset $X^+$,   subscript $sim | div$ indicates whether this coreset is maximally similar ($sim$) or diverse ($div$).

 Why choice of $|C^+|=q |C| = \frac{2}{3} |C| $?  We need $q \in (1/2, 1)$. The larger $q$, the more likely a bad client is included in $C^+$. On the other hand, with a smaller $q$,  good clients may less be utilized and their majority is weaker in the coreset.  We thus heuristically set $q=\frac{2}{3}$. In the worst case where $49\%$ of the clients are bad and all these bad clients happen to be included in $C^+$, we still have $33\%+$ of the clients, who are good, being included in $C^+$. However, this worst case is extremely unlikely. We should expect the majority proportion of good clients to be higher in $C^+$ than that in the original set $C$.

\subsubsection{Aggregation Algorithm}
In each round $t$, the server computes coreset candidates $C^{(t)+}_{sim}$, $C^{(t)+}_{div}$  and averages the gradients in each of these coresets. 
In the case of full client participation $C^{(t)}=C$, we have
\begin{align*}
C^{(t)+}_{sim|div} &= \mathsf{coreset}_{sim|div}( \{\mathbf{w}_i^{(t)}: i \in C \} ; \frac{2}{3}).
\end{align*}
In the case of partial client participation, for efficiency of communication, the server  sends the global solution to a subset of $|C^{(t)}| =o(|C|)$ clients in each round. As such, in each round $t$, the server receives local gradients from only a small subset of responding clients. Let $W^{(t)}$ be the set  of these local gradients plus those last submitted, in round $(t-1)$ or earlier, from the other clients, $C \setminus C^{(t)}$. We then have
\begin{align*}
C^{(t)+}_{sim|div}  &= \mathsf{coreset}_{sim|div}( W^{(t)}   ; \frac{2}{3}).
\end{align*}
Next, for each candidate coreset, $C^{(t)+} \in \{C^{(t)+}_{sim}, C^{(t)+}_{div}\}$, the server obtains a candidate for the global solution by applying arithmetic mean to aggregate,
\begin{align*}
\mathbf{w}^{(t)} = \mathsf{mean} \bigg(  \{\mathbf{w}_i^{(t)}: i \in C^{(t)+} \} \bigg),~\mathbf{x}^{(t)} = \mathbf{x}^{(t-1)} + \eta \mathbf{w}^{(t)}.
\end{align*}
There are thus two global-solution candidates, $\{\mathbf{x}^{(t)}_{sim}, \mathbf{x}^{(t)}_{div} \}$. Both 
 are sent to clients in $C^{(t+1)}$ to start the next round $(t+1)$.

\subsubsection{Client Local Update}
Upon receipt of the two   solution candidates, \{$\mathbf{x}^{(t)}_{sim}$, $\mathbf{x}^{(t)}_{div}$\}, 
each client $i \in C^{(t+1)}$ chooses:  
\[
\mathbf{x}^{(t)} = \arg \max  \{F_i( \mathbf{x}) : \mathbf{x} \in \{\mathbf{x}^{(t)}_{sim}, \mathbf{x}^{(t)}_{div} \}\}
\]
and then apply greedy local update using this $\mathbf{x}^{(t)}$ as usual. 
The rationale behind this heuristic is that the version more valuable according to client $i$'s local criterion should be better representative of the good global solution. 
Note that $F_i(\mathbf{x})$ is computationally expensive as it requires sampling of subsets. Client $i$ can do the following to expedite the choosing: 1) find the subset $S$ corresponding to the $r$ largest coordinates of vector $\mathbf{x}$; and 2) select $\mathbf{x}^{(t)}$ that has the larger value of $f_i(S)$.

%% file: 06_eval.tex
\section{Numerical Results}\label{sec:eval}
We empirically evaluate\footnote{The simulation was conducted thanks to the supercomputing facilities managed by the Research Computing Department at UMASS Boston.}  $\mathsf{RobustFSM}$'s  effectiveness against client attacks. The baseline is 1) $\mathsf{FedCG}$, the original federated algorithm. We also compared to 2) $\mathsf{FedCG\_{median}}$, the variant of $\mathsf{FedCG}$ where Geometric Median is used  for averaging   local solutions; 3) $\mathsf{RobustFSM\_{sim}}$, a variant of $\mathsf{RobustFSM}$ where the server keeps only the coreset candidate using max-similar heuristic; and 4) $\mathsf{RobustFSM\_{div}}$, a variant of $\mathsf{RobustFSM}$ where the server keeps only the coreset candidate using max-diverse heuristic.
\subsection {Experimental Setup}
We assume solving a task of finding top images best representing a collection  of images, under different scenarios.

\textbf{Datasets. } Two real-world datasets are used.
1) CIFAR10 \cite{DBLP:journals/corr/SimonyanZ14a}:  top $r=10$ images from $|D| =$ 60K color images of 10 object categories. Ground set $E$ consists of $500$ images from $D$,   sampled uniformly at random from any category; 
2) PATHMNIST \cite{DBLP:journals/corr/abs-2110-14795}:  top $r=9$ images from $|D| =$ 90K grayscale medical images of 9 categories of pathology tissues. Ground set $E$ consists of $500$ images from $D$, sampled according to a Zipf distribution such that $[30, 75, 60, 10, 20, 90, 10, 10, 195]$ images are selected from the 9 respective categories.

 \textbf{Federated setting. }
There are $n=60$ clients, each hosting a disjoint  sub-collection $D_i$.  The image distribution per client is non-iid according to a Dirichlet process. Each client $i$ has local valuation function $f_i(S) = \sum_{e \in D_i} \max_{s \in S} (1-d(e, s))$ where $d$ is a [0,1]-normalized distance metric between two images $e$ and $s$. $f_i(S)$ is submodular because it is a Facility Location function. When computing the continuous version $\nabla F_i()$ from the discrete version $f_i$, we average over $10$ random subset samples. The learning rate in gradient descend is $\eta = 0.01$.
Regarding metric $d(.)$, for CIFAR10 which consists of color images, we use the bag-of-colors distance  which   is often used to compare  color images \cite{10.1145/2072298.2072034}. For PATHMNIST, whose images are grayscale, we define $d(e, s) = d_1+d_2$ where $d_1$ is the cosine distance between the ResNet18 feature vectors of $e$ and $s$ and $d_2$ is the cosine distance between the feature vector means of the category of $e$ and the category of $s$. 

\textbf{Adversary setting.} We consider $\beta \in \{25\%, 33\%, 49\%\}$ of the clients being malicious, conducting four types of client attack: $\mathsf{Random}$, $\mathsf{Reverse}$, $\mathsf{Include}$, and $\mathsf{Exclude}$. 
In $\mathsf{Include}$  attack, the adversary sets $E_0$ to the $r$ worst-ranked elements of the current global solution set $\mathbf{x}$. In  $\mathsf{Exclude}$ attack, the adversary sets $E_0$ to the $r$ top-ranked elements.

\textbf{Evaluation metrics.} 
The metric for assessing the quality of a global solution $S^*$ is
$ f^* = \frac{1}{|G|} \sum_{i \in G} f_i(S^*)$  where $G$ is the ground-truth subset of good clients. This quantifies how the global solution satisfies an average honest client.  For fair comparison, we normalize $f^* = \frac{f^*-min}{max - min} \in [0, 1]$, where $max$ is the maximum quality which is obtained by running $\mathsf{FedCG}$ under no attack and $min$ is the minimum quality which is obtained by simply choosing a random subset as the final solution (thus, need not run any algorithm).

\subsection{Effect of client attacks on FSM}

\begin{figure}[t]
    \begin{minipage}{0.48\linewidth}
     \includegraphics[width=\textwidth]{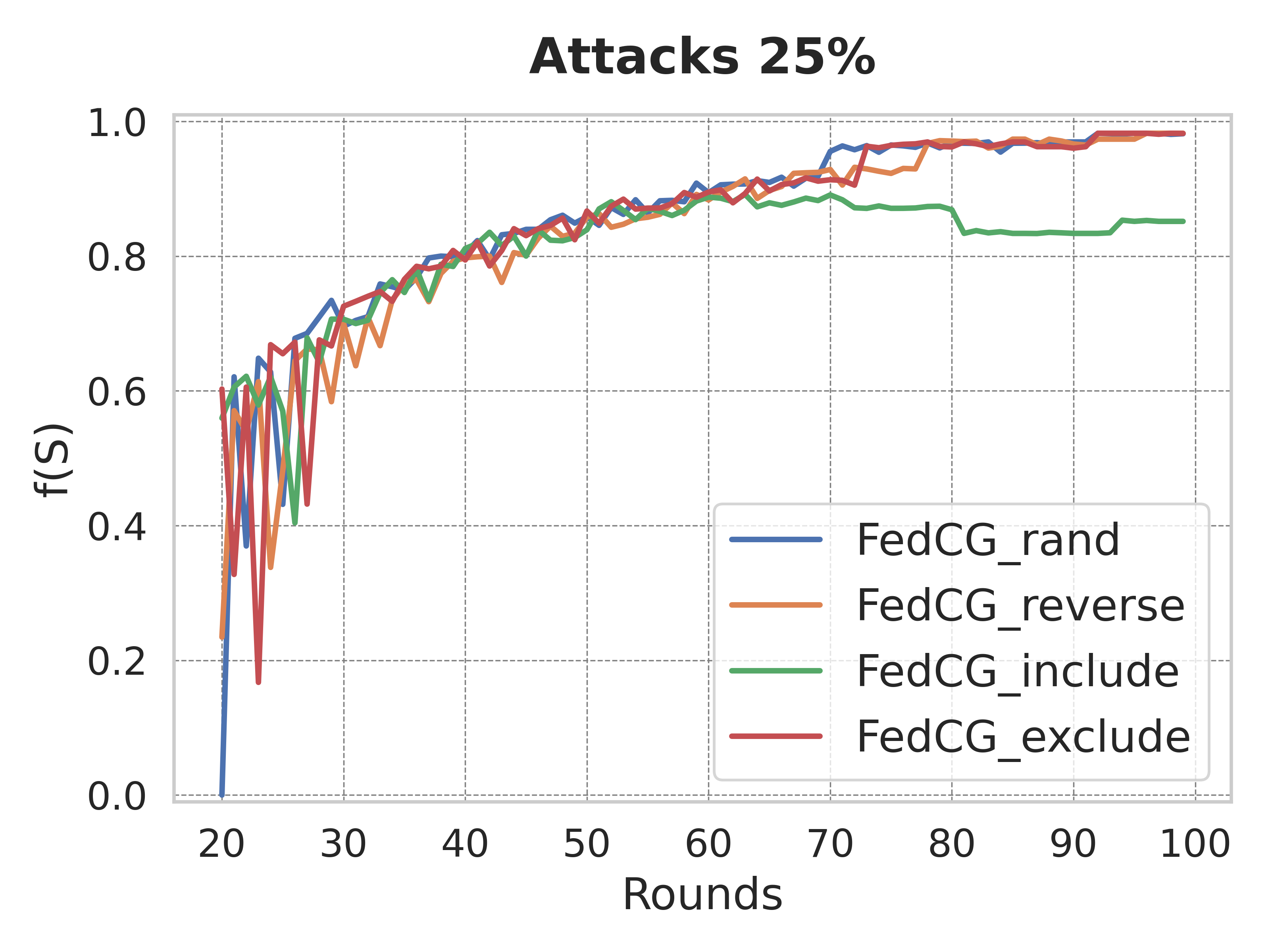}
    \includegraphics[width=\textwidth]{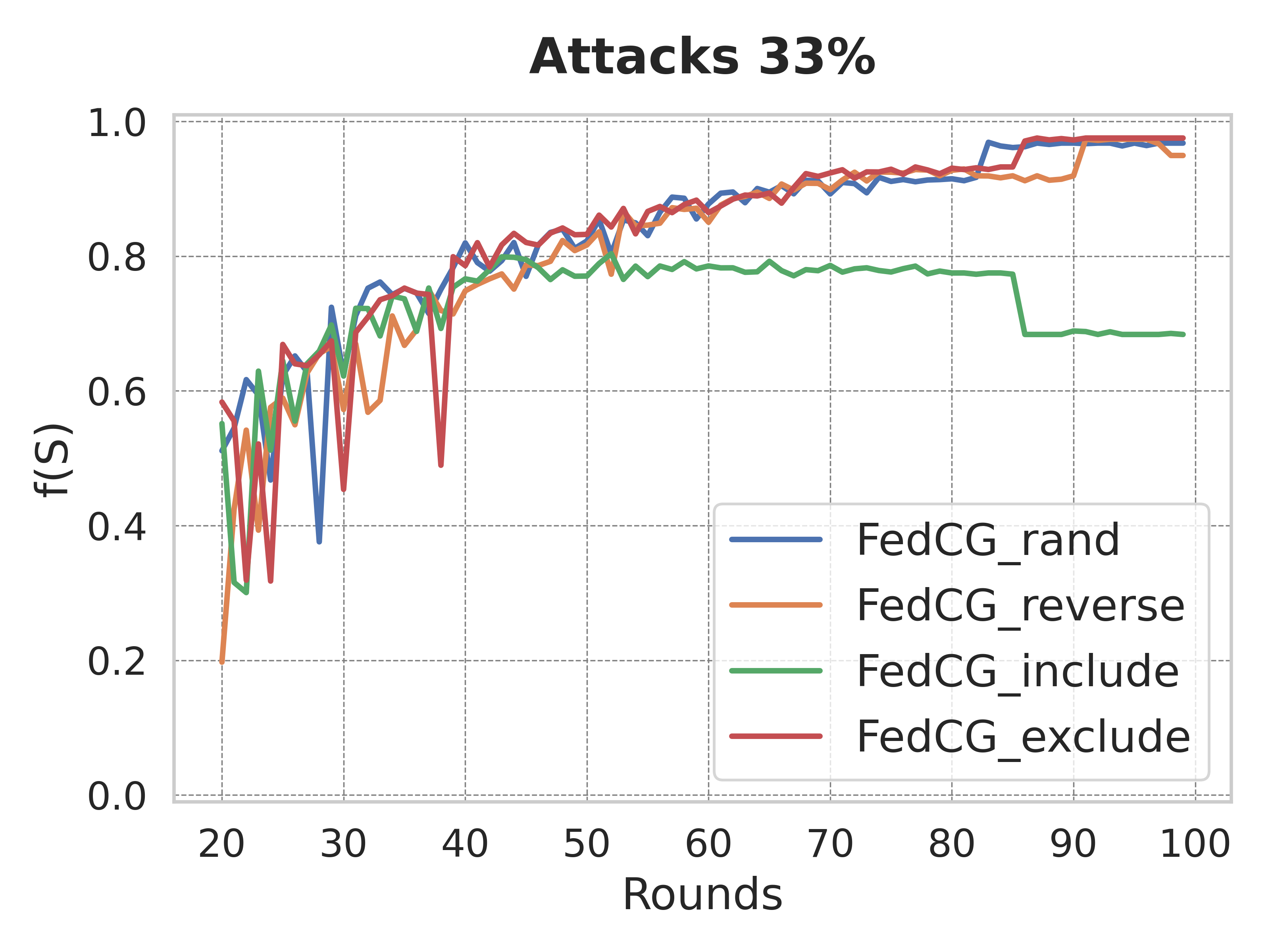}
     \includegraphics[width=\textwidth]{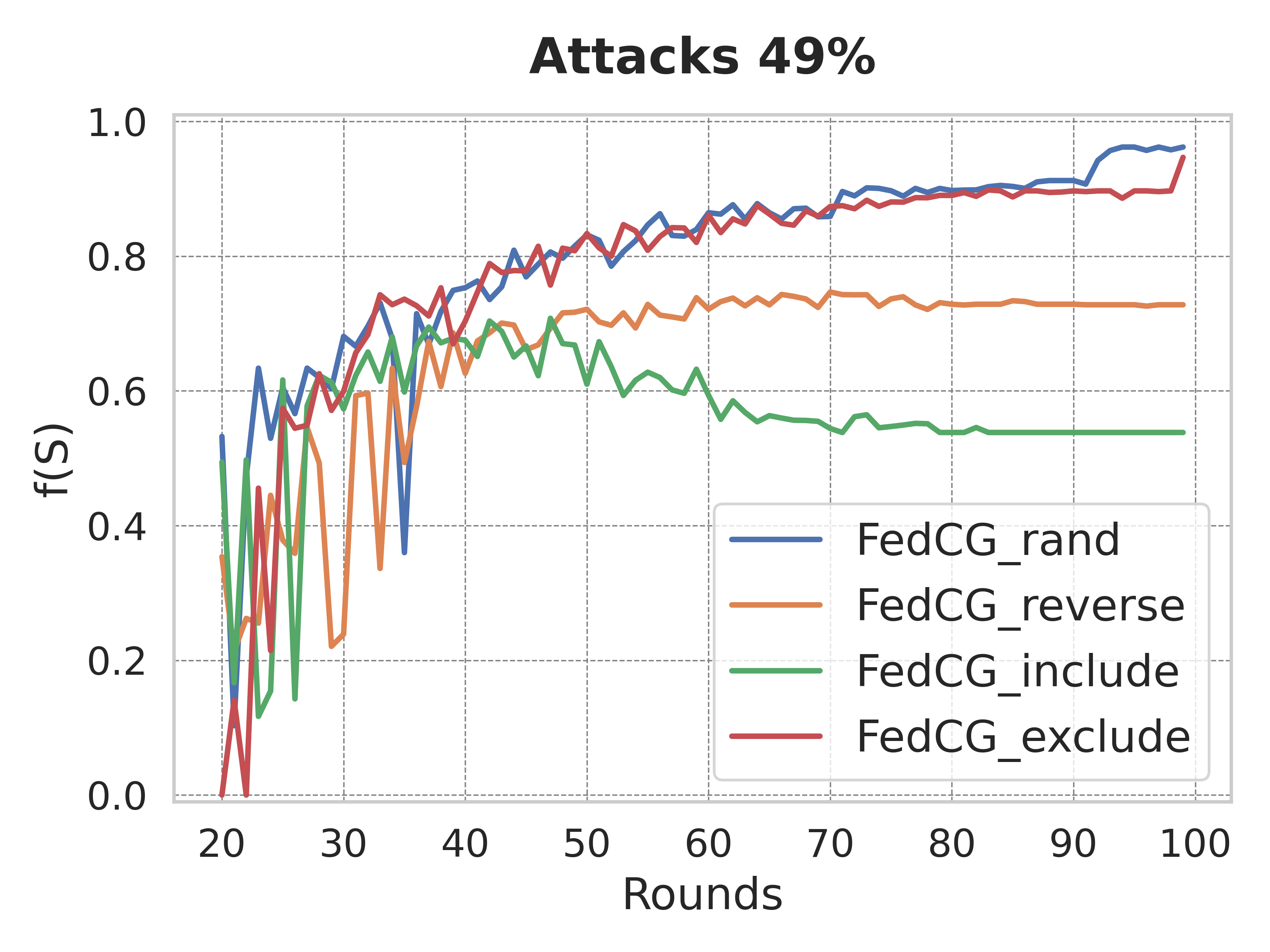}
     \caption*{(a) CIFAR10}
     \label{fig:effect1_cifar10}
    \end{minipage} 
    \hfill
    \begin{minipage}{0.48\linewidth}
        \centering
         \includegraphics[width=\textwidth]{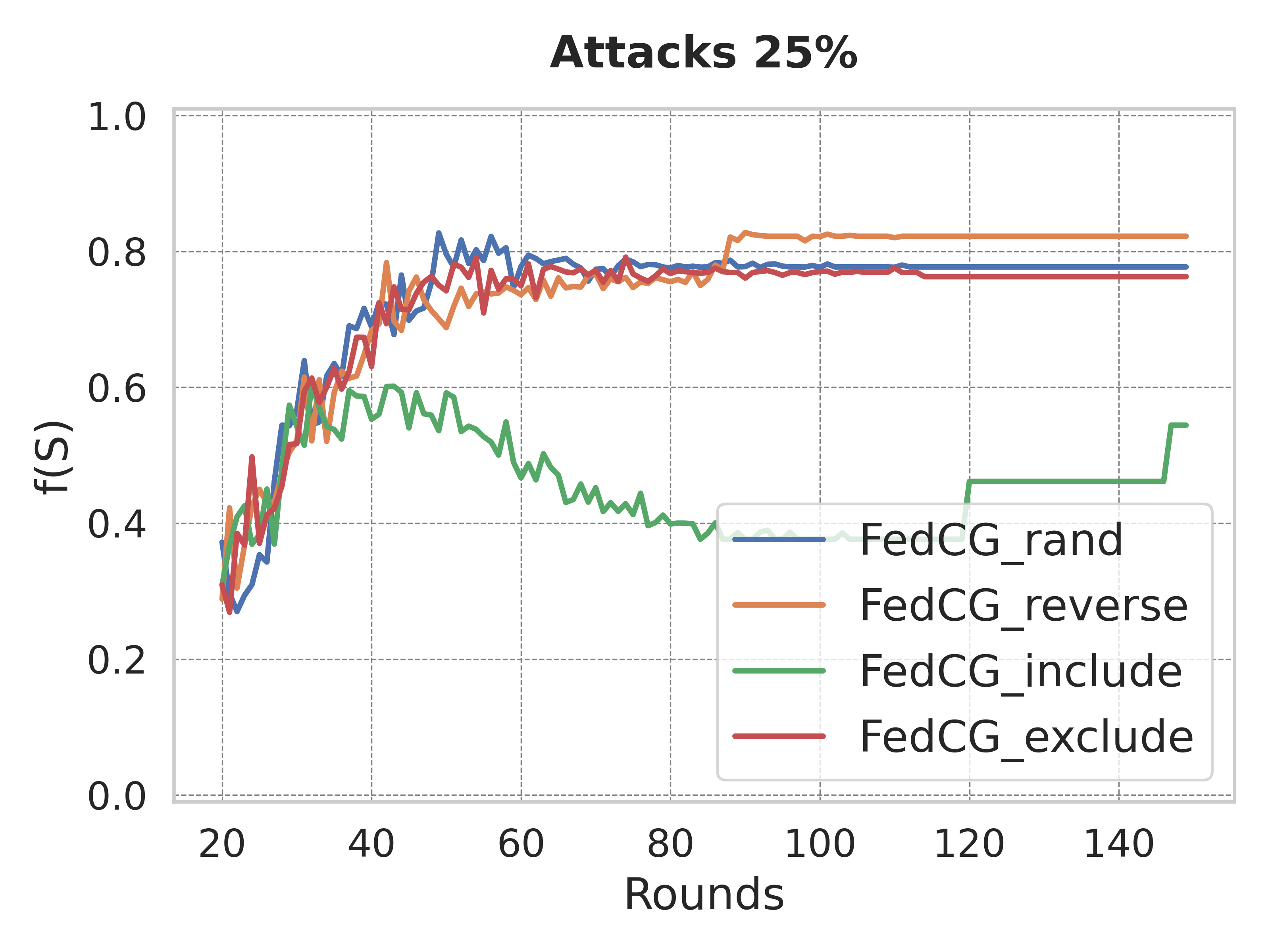}\\
    \includegraphics[width=\textwidth]{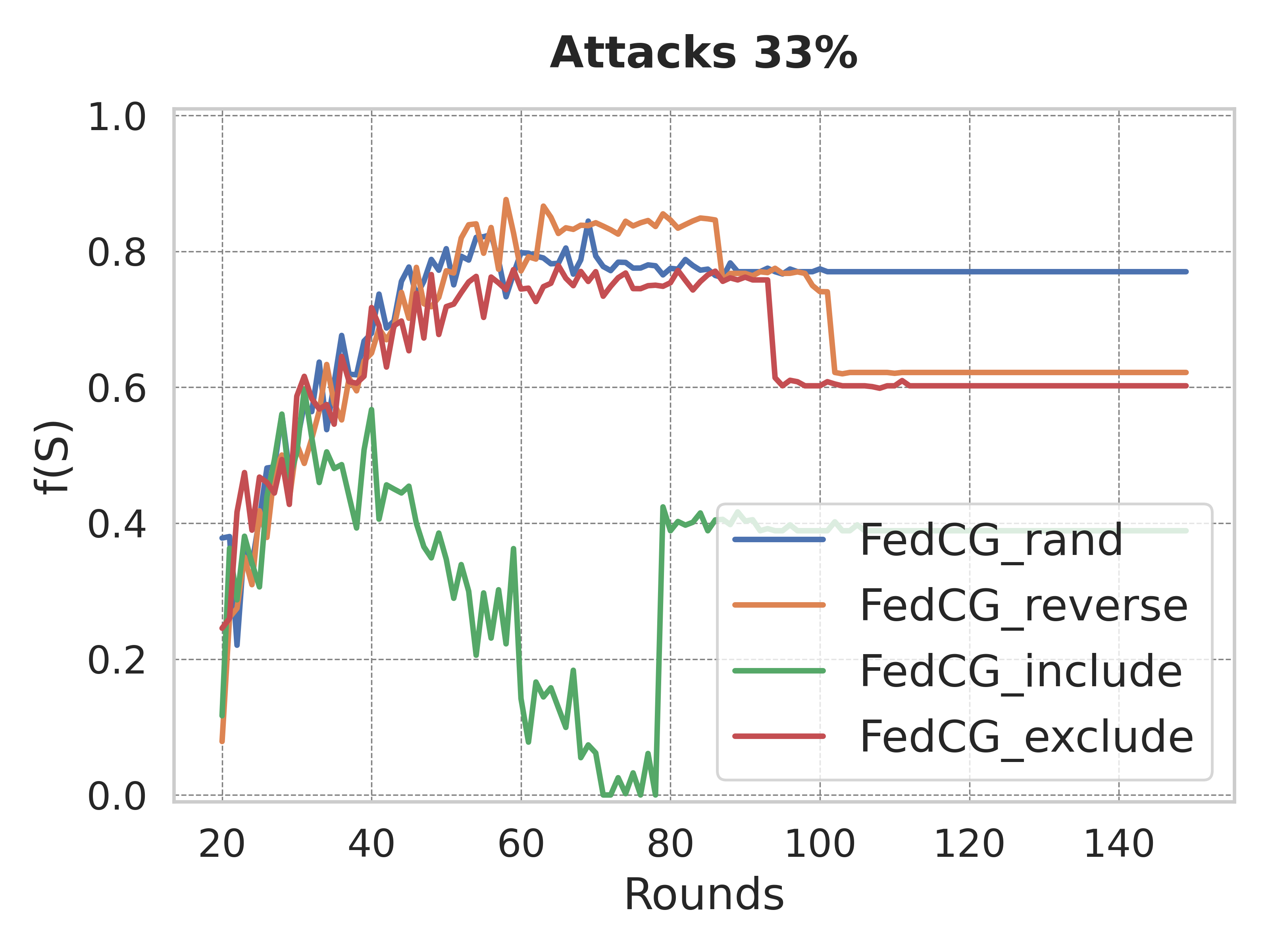}\\
     \includegraphics[width=\textwidth]{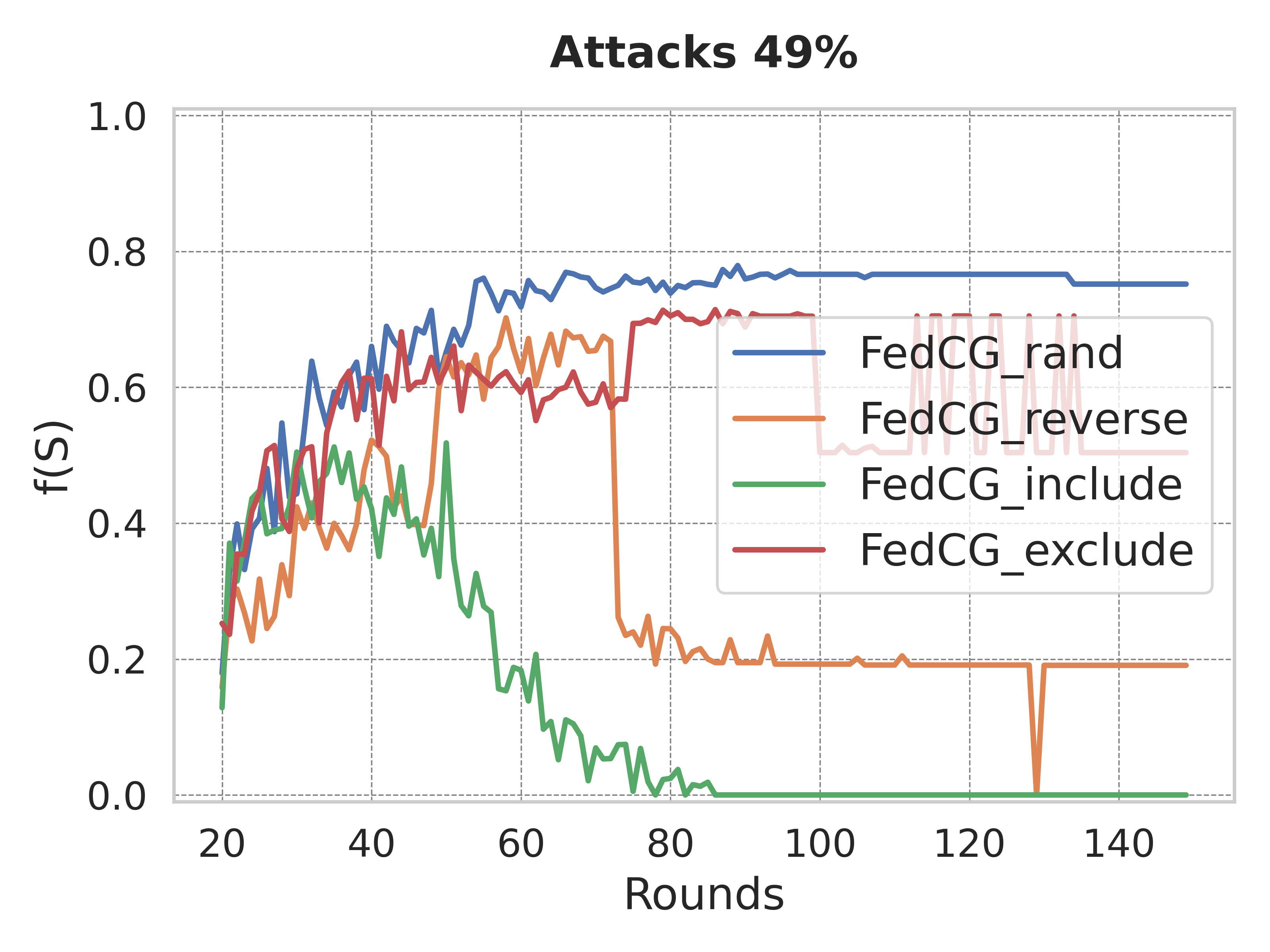}
       \caption*{(b) PATHMNIST}
        \label{fig:effect1_pathmnist}
         \end{minipage}
    
\caption{Effect of different types of attack under different fractions of malicious clients.}
\label{fig:effect1}
\end{figure}

First, we analyze the severity of effect under different types of attack and different degrees of adversity in the conventional federated setting (using the $\mathsf{FedCG}$ algorithm). Four scenarios are considered:  $\mathsf{FedCG}$ under $\mathsf{Random}$ attack ($\mathsf{FedCG\_{rand}}$), $\mathsf{Reverse}$ attack ($\mathsf{FedCG\_{reverse}}$), $\mathsf{Include}$ attack ($\mathsf{FedCG\_{include}}$), and $\mathsf{Exclude}$ attack ($\mathsf{FedCG\_{exclude}}$).
All scenarios converge after 100 federated rounds.  The results for the CIFAR10 dataset are plotted in Figure \ref{fig:effect1}(a). 
It is clearly observed that 1) client attacks result in not only slower convergence rate, but also lower solution quality at convergence, and 2)  the more targeted of the attack, the more severe is its impact. Especially in earlier rounds, client attacks result in very poor solution quality. For example, if the federated process stops after 30 rounds, the resulted quality is only at 60\% - compared to the best case where there is no attack. Although the quality improves with more rounds, which is expected because the majority of the clients is good, the quality at convergence can still be bad depending on the targeted-ness of the attack. Under $\mathsf{Random}$ attack, which is untargeted, the converged quality approaches the best quality ($>90\%$) even if 49\% of the clients is bad. However, under $\mathsf{Reverse}$ attack and $\mathsf{Include}$ attack, which are more targeted, the converged quality is below 90\% (with 33\% clients being bad) and below 70\% (with 49\% clients being bad).

Figure \ref{fig:effect1}(b) demonstrates the impact of attacks for the PATHMNIST dataset. The same pattern applies, i.e., attacks are more impactful if more targeted, but the degree of damage is much more severe. Under $\mathsf{Include}$ attack, the quality is below 50\%  when only 25\% of the clients is bad. With 33\% bad clients, the quality under $\mathsf{Reverse}$ or $\mathsf{Exclude}$ is around 60\%, while the quality under $\mathsf{Include}$ attack is 40\%. With 49\% bad clients, the $\mathsf{Include}$ brings the quality down to 0\%. This simulation study proves that the quality damage due to client attacks can be substantial and thus must be addressed.

\subsection{Improvement of $\mathsf{RobustFSM}$ over $\mathsf{FedCG}$}

\begin{figure}[t]
    \begin{minipage}{0.48\linewidth}
     \includegraphics[width=\textwidth]{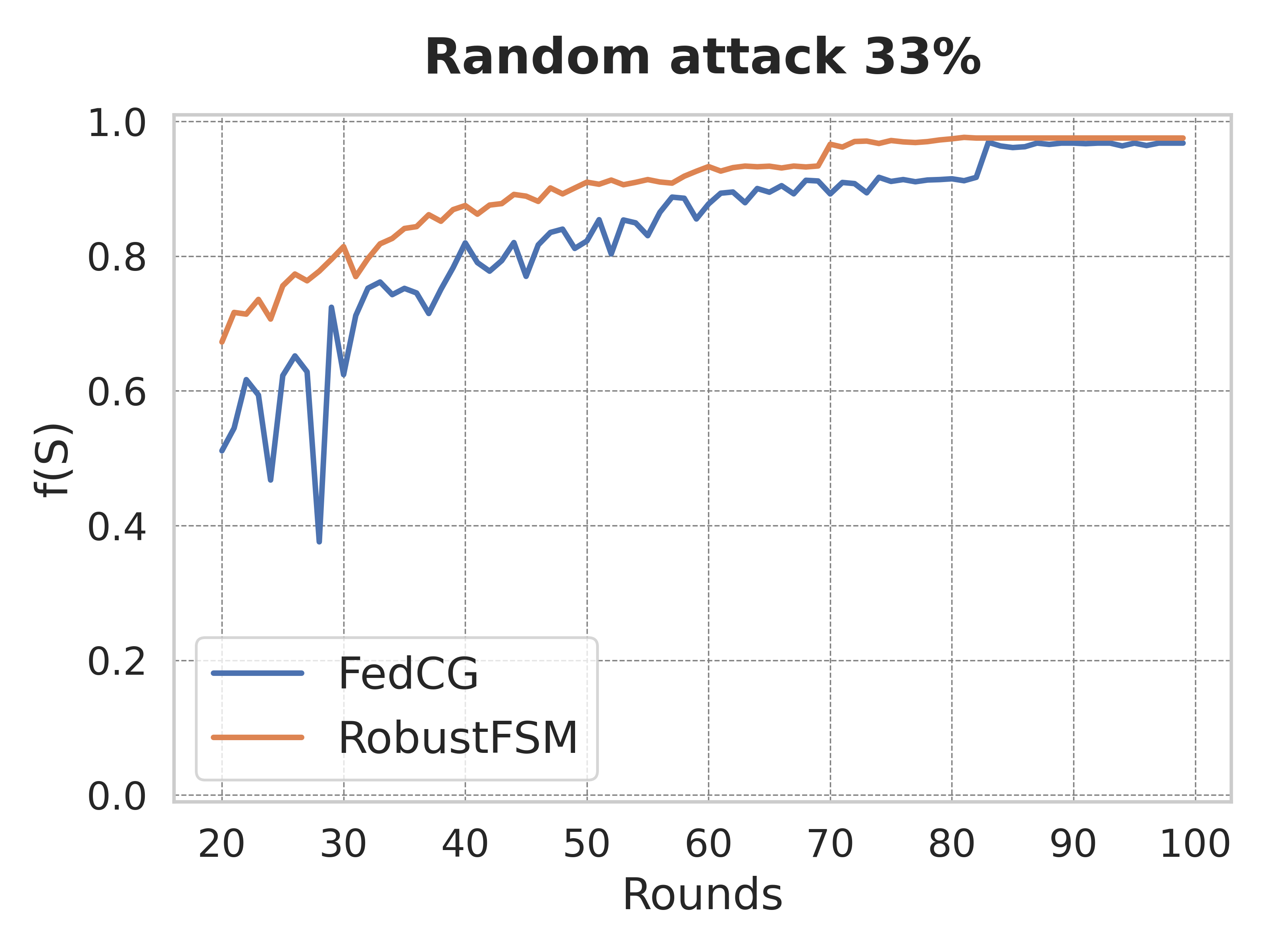}
    \includegraphics[width=\textwidth]{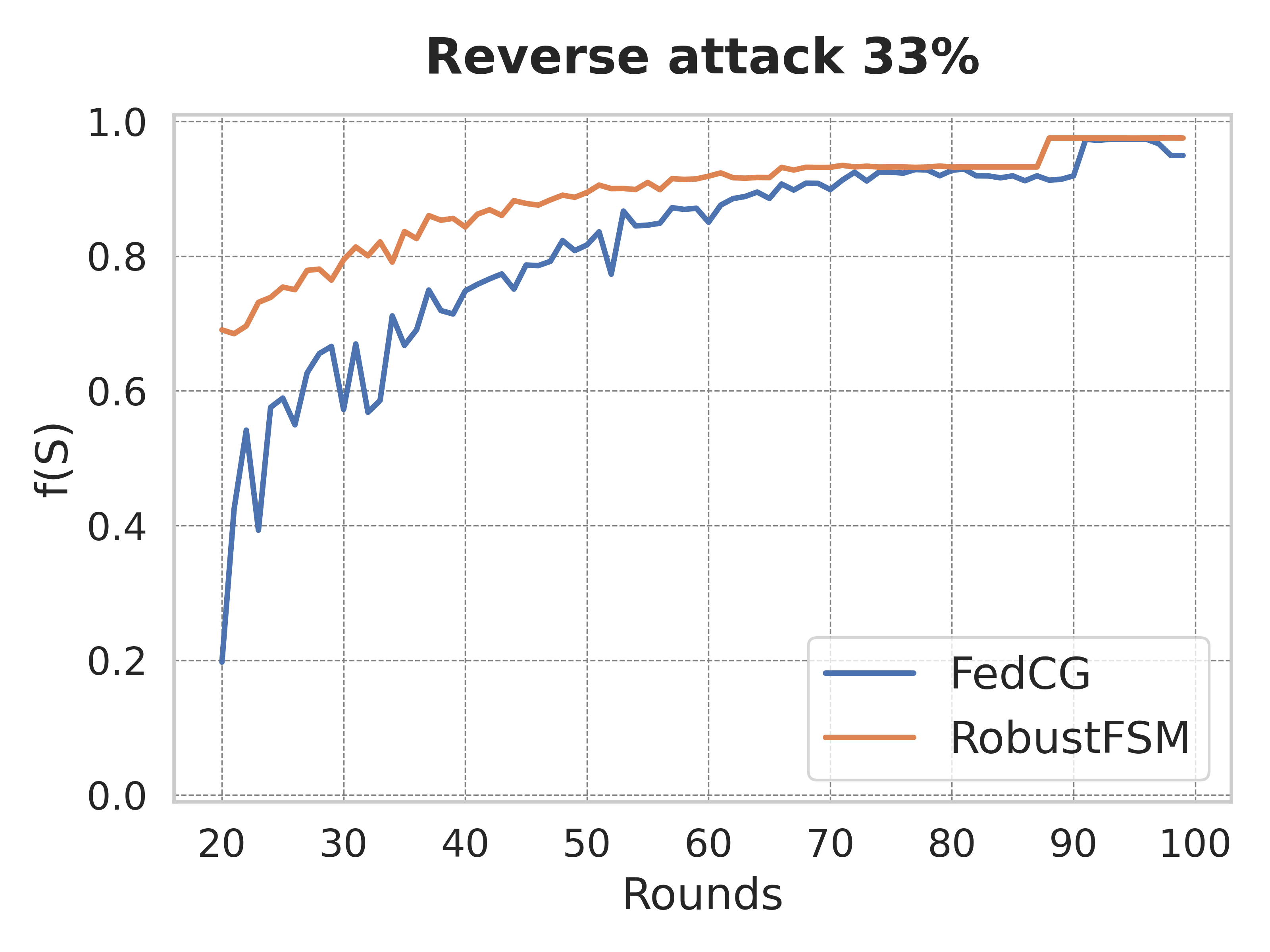}
        \includegraphics[width=\textwidth]{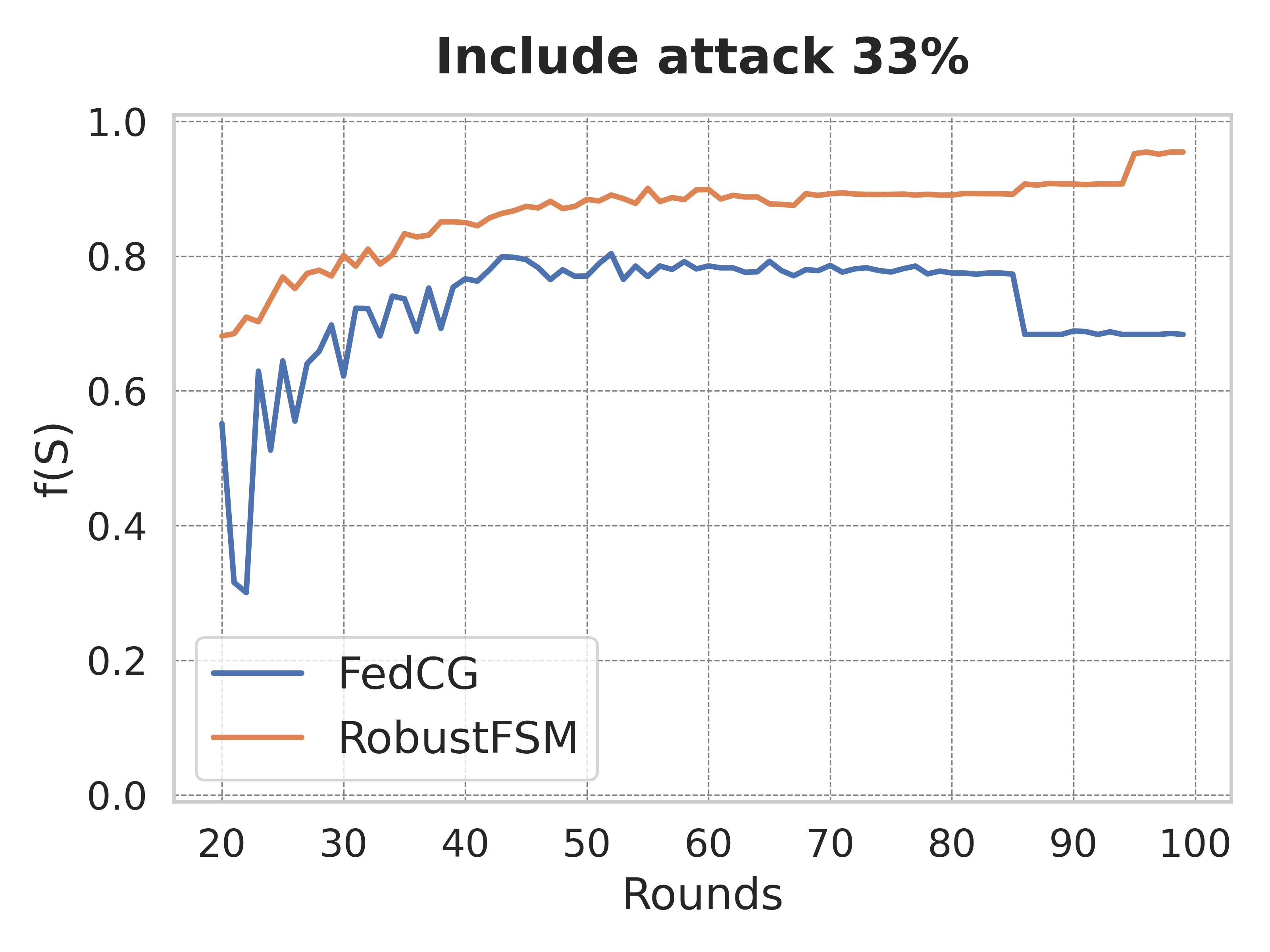}
   \caption*{(a) CIFAR10}
     \label{fig:compare1_cifar10}
    \end{minipage} 
    \hfill
    \begin{minipage}{0.48\linewidth}
        \centering
        \includegraphics[width=\textwidth]{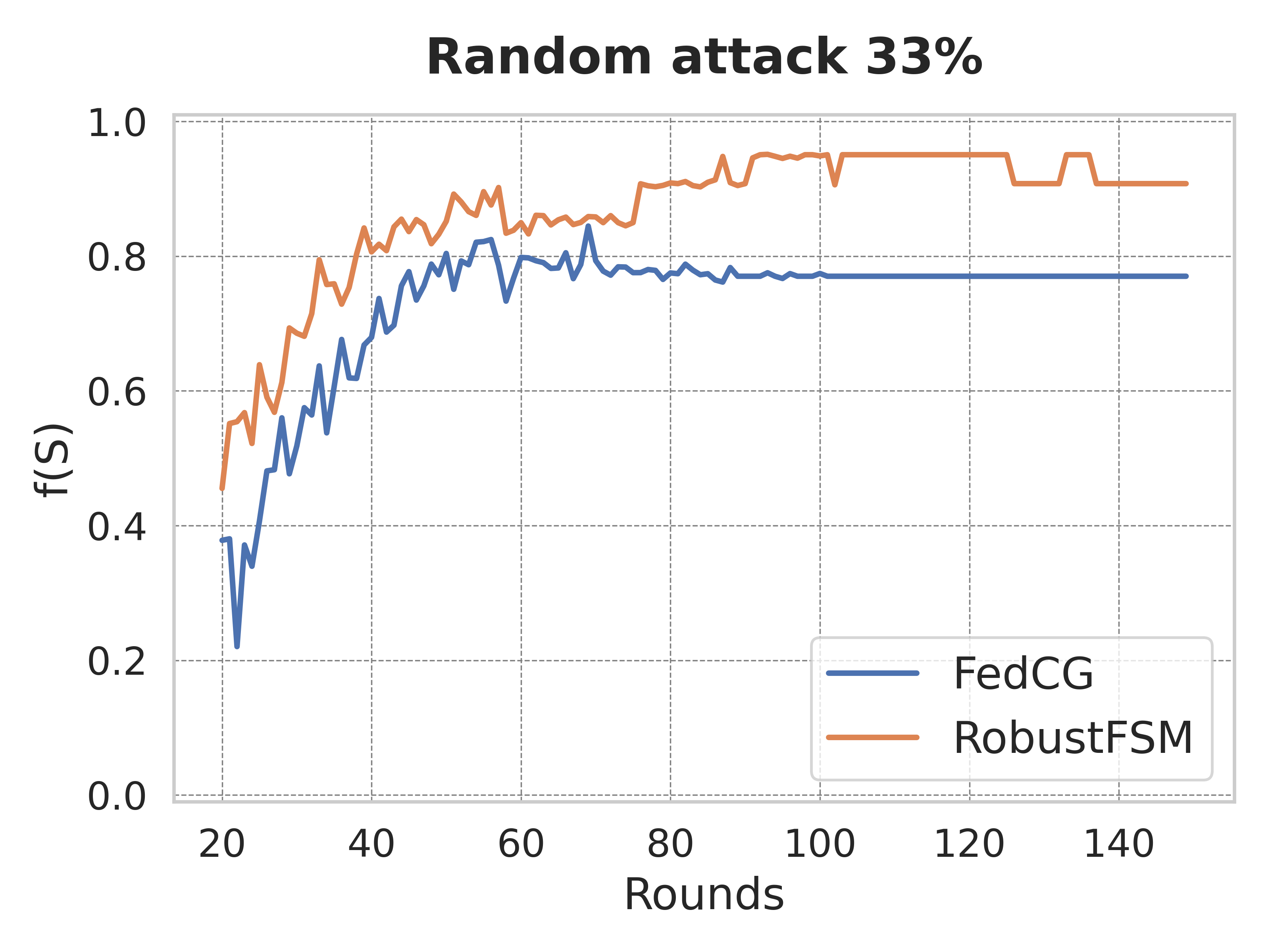}
    \includegraphics[width=\textwidth]{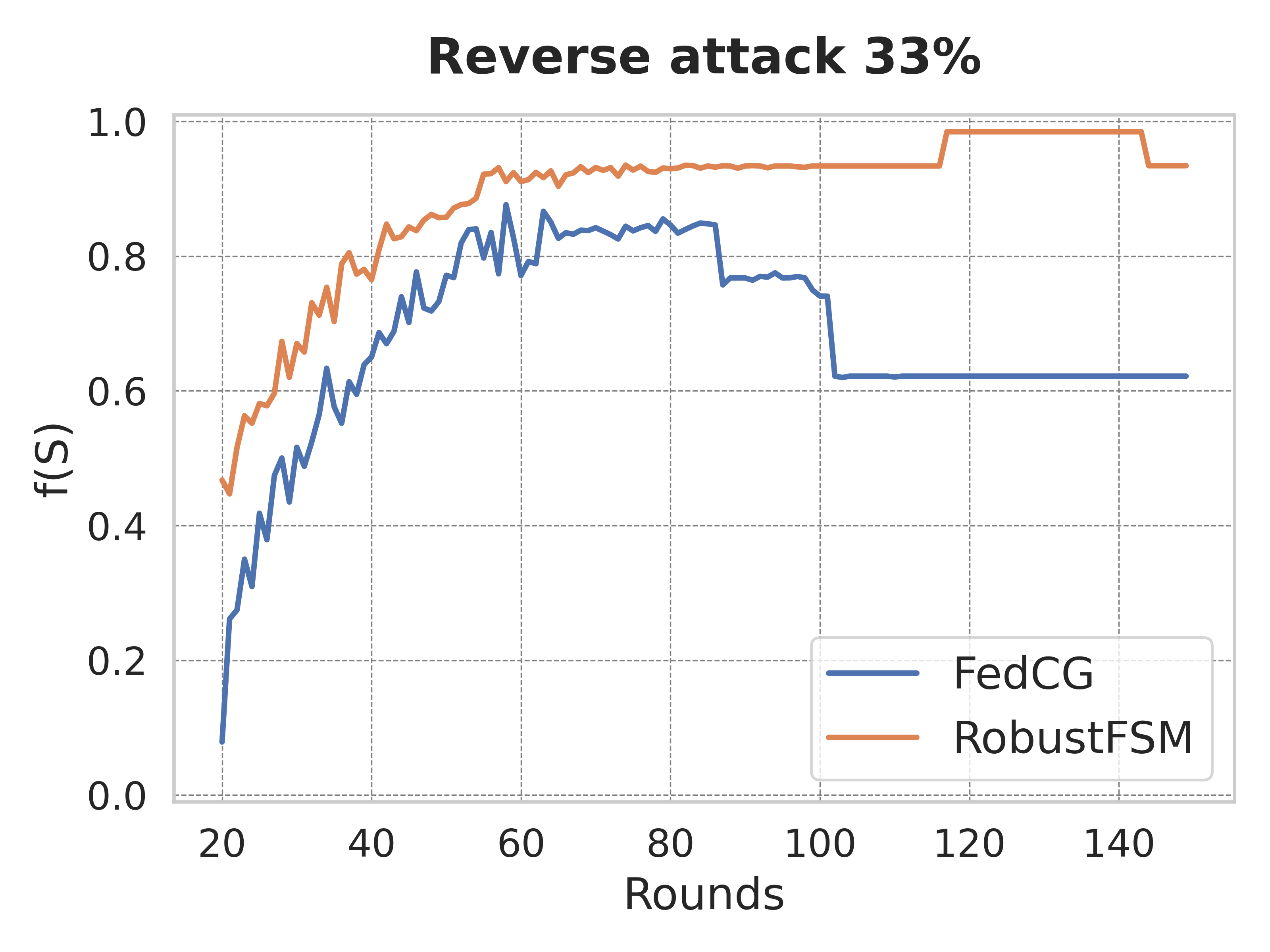}
        \includegraphics[width=\textwidth]{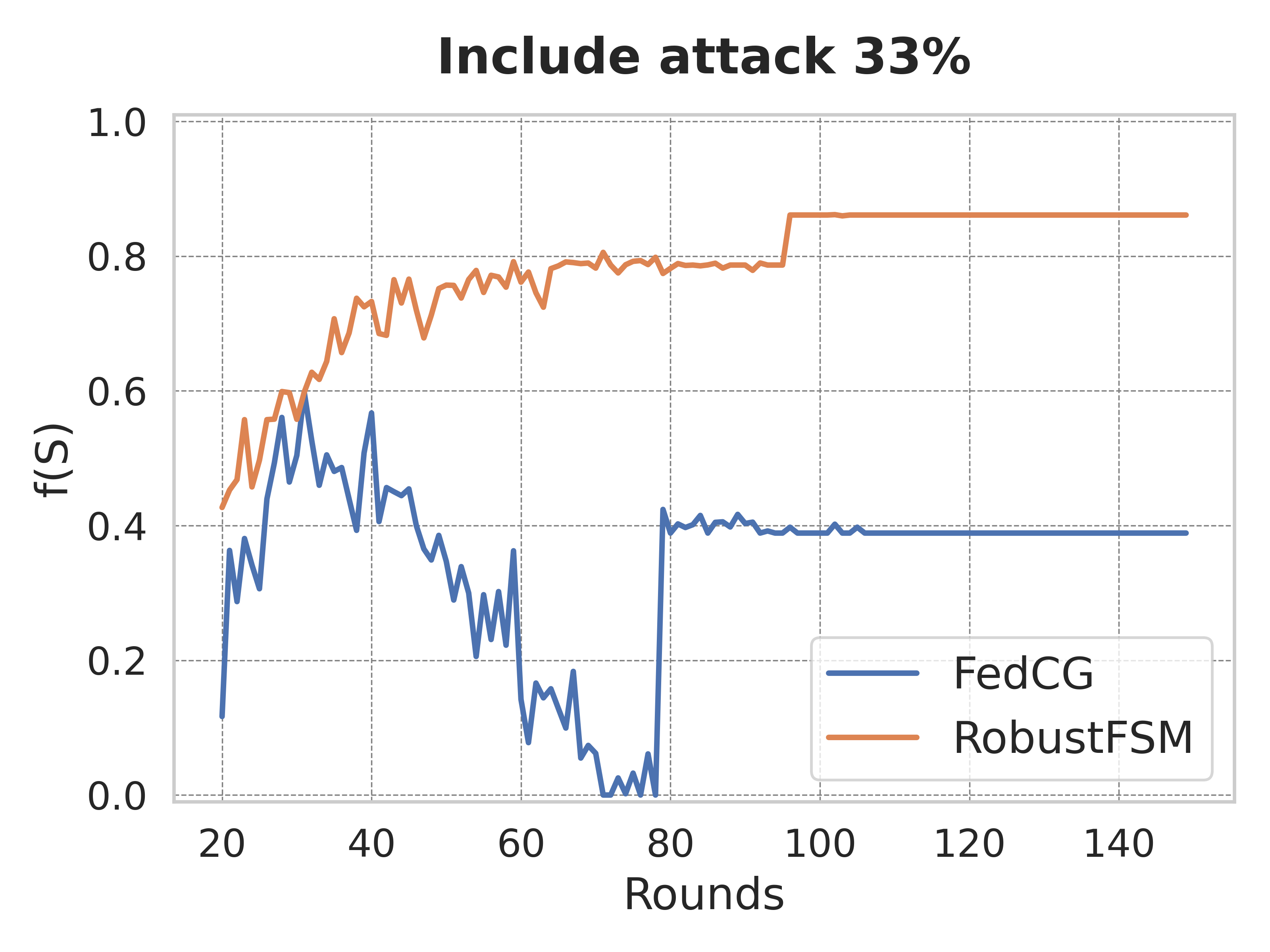}
     \caption*{(b) PATHMNIST}
        \label{fig:compare1_pathmnist}
         \end{minipage}
    
\caption{Comparison of $\mathsf{RobustFSM}$ versus $\mathsf{FedCG}$  under different attack types, where  33\% of clients is bad.}
\label{fig:compare1}
\end{figure}

Next, we go into each type of attack and analyze how  $\mathsf{RobustFSM}$ can be more robust  versus $\mathsf{FedCG}$.  For the sake of illustration, we discuss the case of 33\% clients being bad; similar observations hold for the 25\% and 49\% cases. Let us first focus on the CIFAR10 dataset. As observed in   Figure \ref{fig:compare1}(a), it is consistent in all scenarios that   $\mathsf{RobustFSM}$ is superior to $\mathsf{FedCG}$  in terms of both convergence rate and better quality. Specifically,  $\mathsf{RobustFSM}$ offers a quality above 95\% at convergence under any attack. For example, under $\mathsf{Include}$ attack - the most severe attack,  $\mathsf{RobustFSM}$ converges with a quality value of 96\% compared to 80\% of  $\mathsf{FedCG}$; this is a  20\% quality improvement.  It is noted that the quality improvement of $\mathsf{RobustFSM}$ takes effect even in early rounds of the federated process.

\begin{figure*}[t]
 \begin{minipage}{0.65\linewidth}
        \subfigure[CIFAR10: top-10 images]{
        		\includegraphics[width= \textwidth]{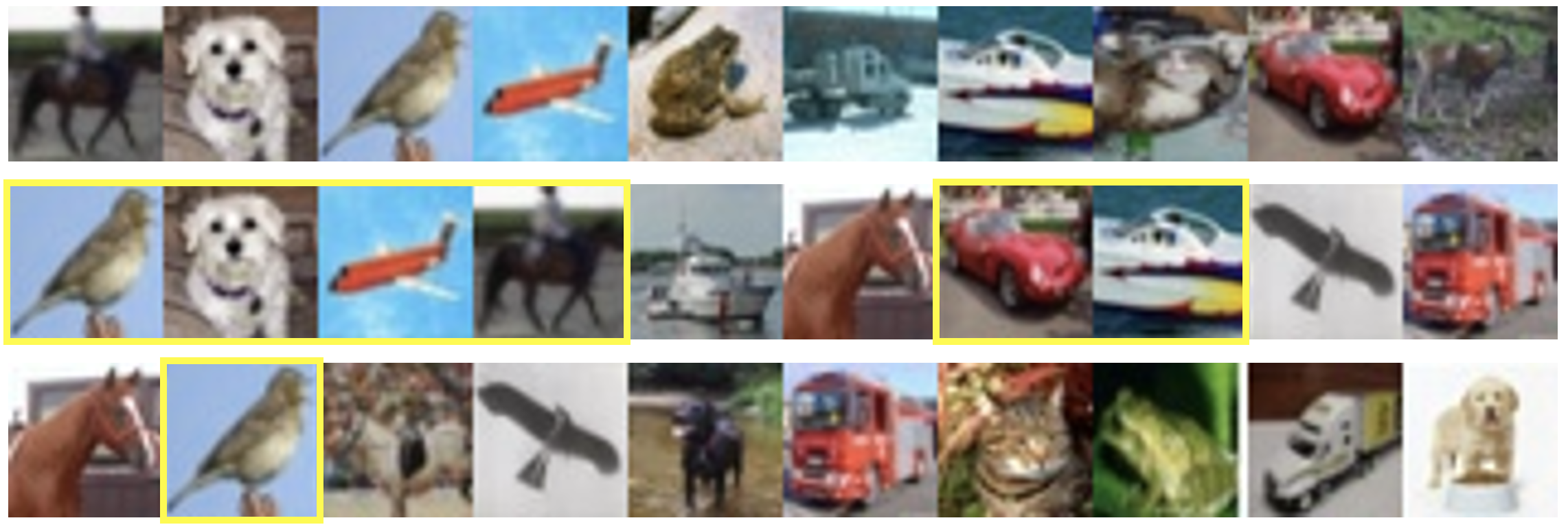} 
        }
        \subfigure[PATHMNIST: top-9 images]{
	        \includegraphics[width= \textwidth]{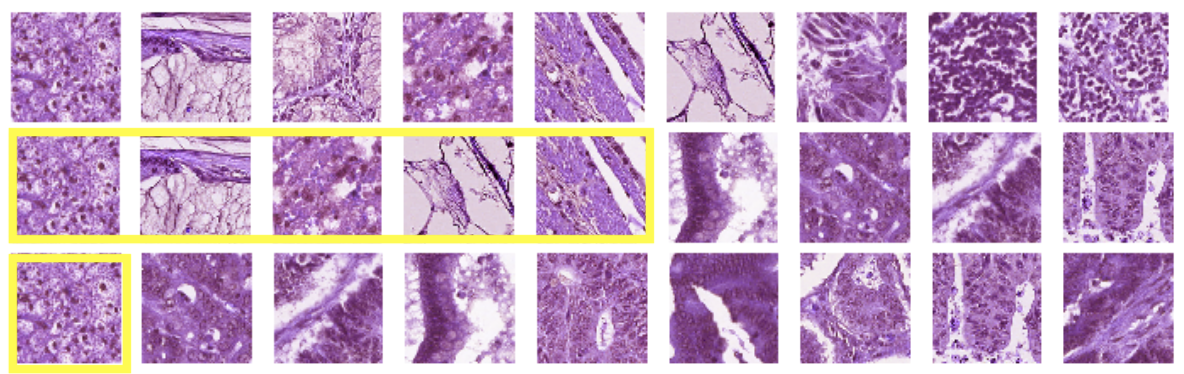} 
       }
       \caption{Visual comparison of top representative images. In each subfigure (a) or (b):  (top row) $\mathsf{FedCG}$ under no attack, (middle row) $\mathsf{RobustFSM}$ under 33\% $\mathsf{Include}$ attack, and (bottom row) $\mathsf{FedCG}$ under 33\% $\mathsf{Include}$ attack. Yellow-marked images are those that match precisely with the no-attack solution. }
     \label{fig:visual_compare}
     \end{minipage}
     \hfill
    \begin{minipage}{0.32\linewidth}
    \centering
    \includegraphics[width=0.8\textwidth]{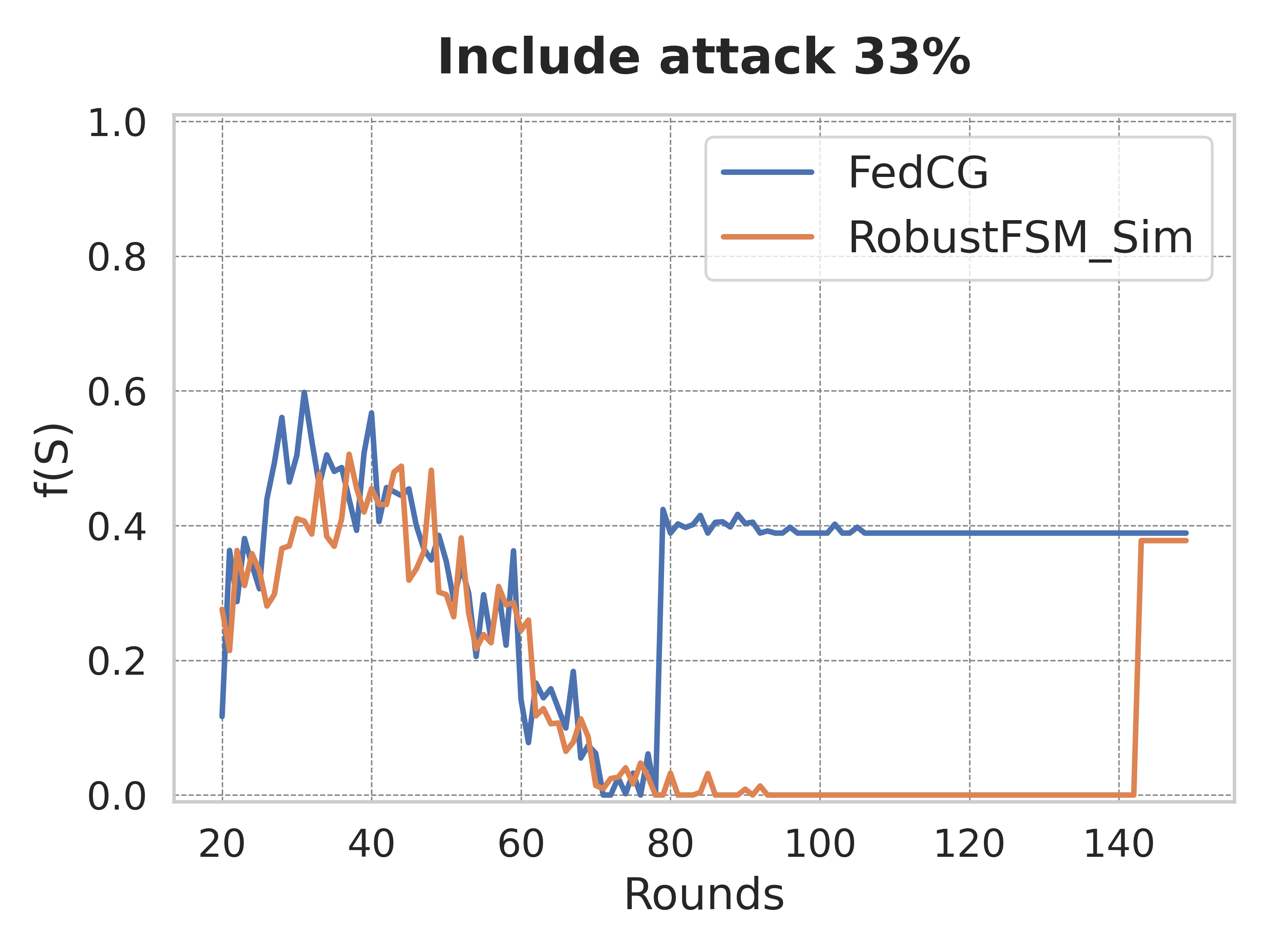}
    \includegraphics[width=0.8\textwidth]{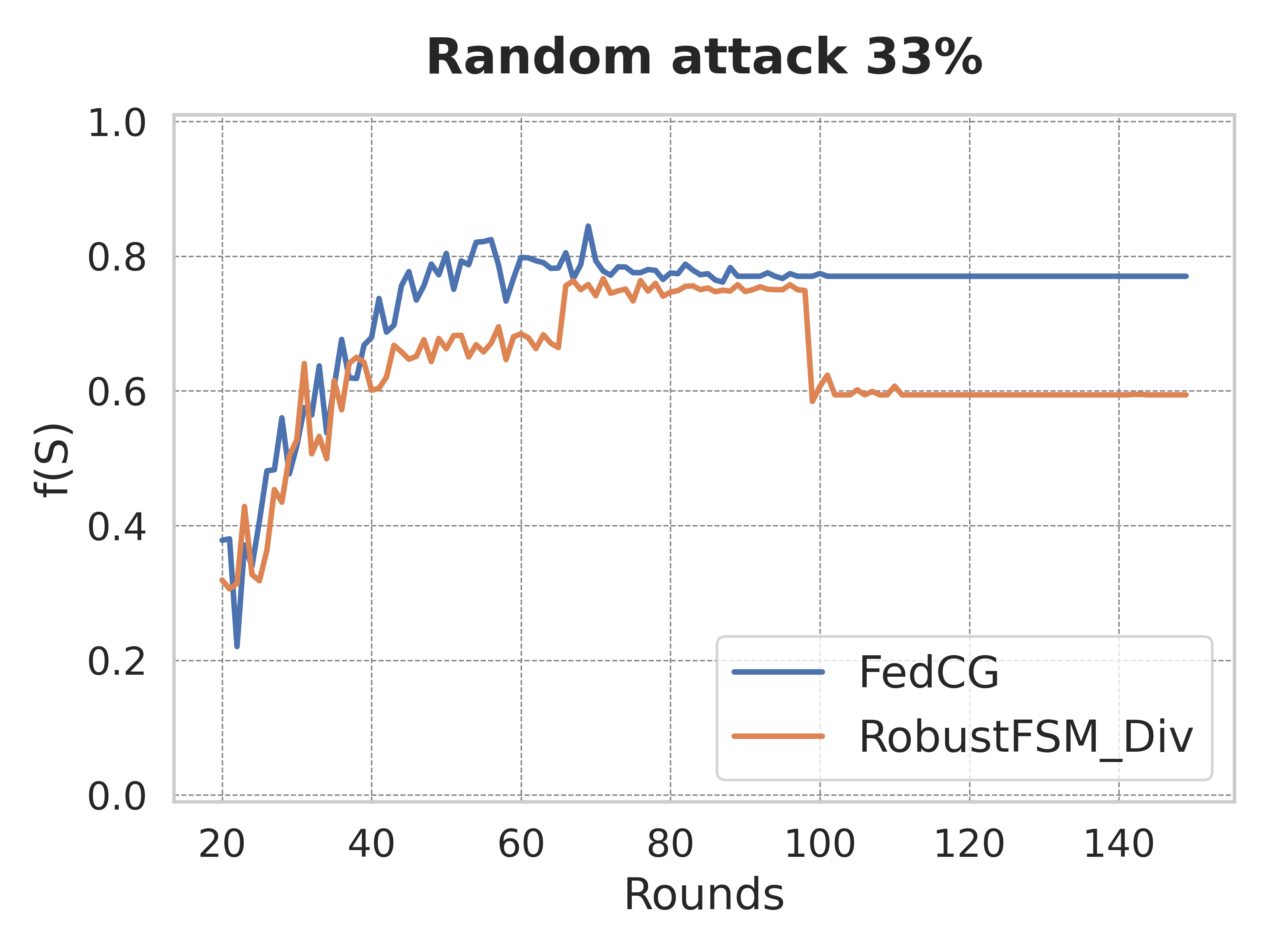}
    \includegraphics[width=0.8\textwidth]{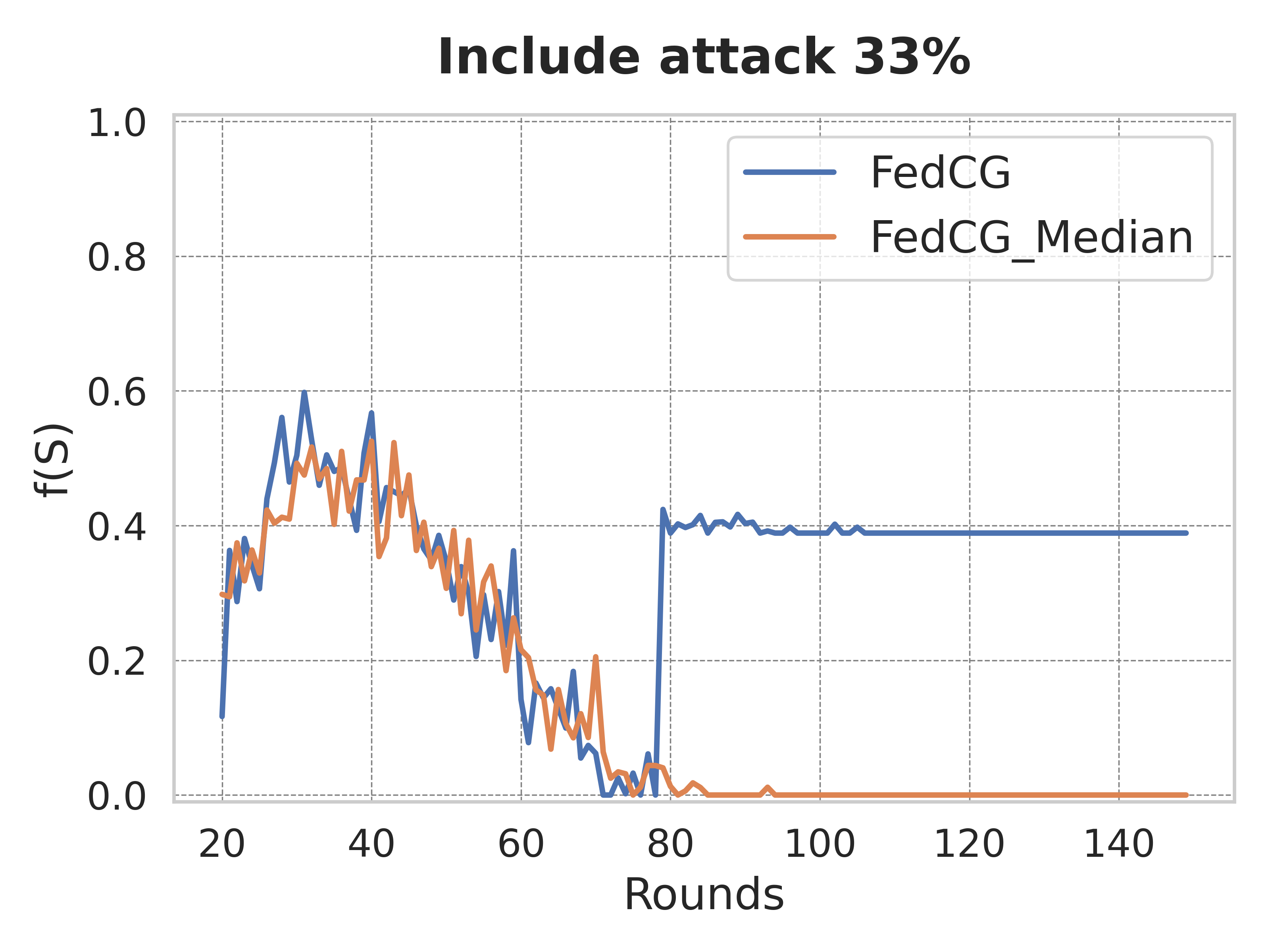}
     \caption{Max-similarity-only or  max-diversity-only heuristic and geometric median are bad  under certain attacks.}
     \label{fig:bad_evidence}
    \end{minipage} 
\end{figure*}

Figure \ref{fig:visual_compare}(a) provides a visual comparison of the top-10 representative images under 33\% $\mathsf{Include}$ attack of $\mathsf{RobustFSM}$ and $\mathsf{FedCG}$. Here, we should expect the 10 final images to represent the 10 different categories equally. In the visualization, under no attack, $\mathsf{FedCG}$ effectively results in 10 images each belonging to a different category. With 33\% $\mathsf{Include}$ attack, we can see that 1) $\mathsf{RobustFSM}$ returns 6 exact same images, including the same order for the top-4, and 2) there are three categories, \{horse,  ship, and bird\}, that have a redundant image (those marked with the red color in the figure). In comparison, $\mathsf{FedCG}$ has only one identical image (the bird) and four redundant image categories, \{horse, bird,  truck,  dog\}. Furthermore, its top-4 images include only two correct-category images, \{horse, bird\}.

Now consider  PATHMNIST. As discussed earlier and seen in Figure \ref{fig:effect1}(b), for this dataset, $\mathsf{FedCG}$ is more severely affected by client attacks. However, $\mathsf{RobustFSM}$ is still highly robust. Figure \ref{fig:compare1}(b) demonstrates how it improves over $\mathsf{FedCG}$ when 33\% of the clients is bad. Under 
$\mathsf{Random}$ and $\mathsf{Reverse}$ attacks, the quality can converge close to 100\%. In the worst case of attack, the $\mathsf{Include}$ attack, we can achieve 85\%,  more than 2x better than  the 40\% quality of $\mathsf{FedCG}$. Not only that, $\mathsf{RobustFSM}$  quickly reaches 80\% quality after early rounds (40+ rounds). Figure \ref{fig:visual_compare}(b) provides a visualization of the top-9 images, comparing $\mathsf{RobustFSM}$ and $\mathsf{FedCG}$ under the same 33\% $\mathsf{Include}$  attack. The former returns 5 exact images out of 9, whereas $\mathsf{FedCG}$ returns only 1 exact image.

\subsection{Justification for $\mathsf{RobustFSM}$}

Figure \ref{fig:bad_evidence} provides empirical evidence to justify our strategy of  simultaneously considering two versions for the global solution: one version based on the max-similar coreset $C^+_{sim}$ and one version based on the max-diverse coreset  $C^+_{div}$, instead of relying on only one exclusive criterion for the coreset. Here, for the sake of illustration, we plot the results for   PATHMNIST. As seen,   if we change our algorithm such that  only $C^+_{sim}$ is used (the $\mathsf{RobustFSM\_{sim}}$ algorithm), the quality is worse than $\mathsf{FedCG}$ in  the case of $\mathsf{Include}$ attack (Figure \ref{fig:bad_evidence}(a)).
 This is explainable. In $\mathsf{Include}$ attack, the bad clients tend to have similar local solutions because they are intended to include certain elements. As such, if we applied the max-similar heuristic to detect good clients, we would mistakenly include the bad clients (who are  similar). Now, if we change our algorithm to use    only $C^+_{div}$ (the $\mathsf{RobustFSM\_{div}}$ algorithm), the result is worse than $\mathsf{FedCG}$ in the case of $\mathsf{Random}$ attack (Figure \ref{fig:bad_evidence}(b)). This is explainable too. In this untargeted attack, good clients tend to be more similar to each other, compared to the group of bad clients who are more diverse due to lack of attack coordination.  As such, if we applied the max-diverse heuristic to detect good clients, we would mistakenly include the bad clients (who are diverse).

Next, Figure \ref{fig:bad_evidence}(c) shows an evidence that Geometric Median is not a good averaging aggregator for FSM;  although it is often used as a robust aggregator in machine learning. If we change $\mathsf{FedCG}$ such that Geometric Median, not Arithmetic Mean, is used to average local solutions (the $\mathsf{RobustFSM\_{median}}$ algorithm), the quality is substantially worse in the case of $\mathsf{Include}$ attack. The above results substantiate the point we raised in earlier the paper that FSM is a very different optimization problem compared to Federated Learning, and as such requires new investigations into the robustness problem.

%% file: 07_conclusion.tex
\section{Conclusions}\label{sec:conclusion}

This paper addresses the problem of federated submodular maximization (FSM) 
with malicious clients. We have investigated several client attacks and their severity. 
These attacks are unlike those often seen in Federated Learning, thus necessitating 
new heuristics and mitigating methods for FSM. We have proposed an effective robust algorithm, 
$\mathsf{RobustFSM}$, which can substantially improve over today's FSM standard, especially 
when attacks are severe. As the robustness problem in FSM is still early in research, 
our findings are useful as a new benchmark for comparison. In future work, we will further 
improve the quality solution of $\mathsf{RobustFSM}$ and extend its robustness against 
more types of client attacks.